\documentclass{article}

\usepackage{PRIMEarxiv}

\usepackage[utf8]{inputenc} 
\usepackage[T1]{fontenc}    
\usepackage{hyperref}       
\usepackage{url}            
\usepackage{booktabs}       
\usepackage{amsfonts}       
\usepackage{nicefrac}       
\usepackage{microtype}      
\usepackage{lipsum}
\usepackage{fancyhdr}       
\usepackage{graphicx}  
\usepackage{textcomp}
\usepackage{subfigure}
\usepackage{url}
\usepackage{verbatim}
\usepackage{graphicx}
\usepackage{amssymb}
\usepackage{amsmath}
\usepackage{array,booktabs}
\usepackage{multirow}
\usepackage{bm}
\usepackage{bbding}
\usepackage{pifont}
\usepackage{makecell}
\usepackage{xcolor}
\usepackage{array}
\usepackage{tabularx}
\usepackage{newfloat}
\usepackage{listings}
\usepackage{algorithm}
\usepackage{algorithmic}
\usepackage{hyperref}
\graphicspath{{media/}}     

\title{GPT-4 Enhanced Multimodal Grounding for Autonomous Driving: Leveraging Cross-Modal Attention with Large Language Models}
\author{Haicheng Liao\thanks{Both authors contributed equally to this research; \dag Corresponding author.}\\
University of Macau\\
Macau SAR, China\\
{\tt\small yc27979@um.edu.mo}\\
\and \textbf{Huanming Shen}$^{*}$\\
UESTC\\
Chengdu, China\\
{\tt\small HuanmingShen@outlook.com}\\
\and
\textbf{Zhenning Li$^{\dag}$}\\
University of Macau\\
Macau SAR, China\\
{\tt\small zhenningli@um.edu.mo}
\and
\textbf{Chengyue Wang}\\
University of Macau\\
Macau SAR, China\\
{\tt\small chengyuewang@um.edu.mo}
\and
\textbf{Guofa Li}\\
Chongqing University\\
Chongqing, China\\
{\tt\small liguofa@cqu.edu.cn}
\and
\textbf{Yiming Bie}\\
Jilin University\\
Jilin, China\\
{\tt\small yimingbie@126.com}
\and
\textbf{Chengzhong Xu$^{\dag}$}\\
University of Macau\\
Macau SAR, China\\
{\tt\small czxu@um.edu.mo}
}

\begin{document}
\maketitle

\begin{abstract}
In the field of autonomous vehicles (AVs), accurately discerning commander intent and executing linguistic commands within a visual context presents a significant challenge. This paper introduces a sophisticated encoder-decoder framework, developed to address visual grounding in AVs. Our Context-Aware Visual Grounding (CAVG) model is an advanced system that integrates five core encoders—Text, Image, Context, and Cross-Modal—with a Multimodal decoder. This integration enables the CAVG model to adeptly capture contextual semantics and to learn human emotional features, augmented by state-of-the-art Large Language Models (LLMs) including GPT-4. The architecture of CAVG is reinforced by the implementation of multi-head cross-modal attention mechanisms and a Region-Specific Dynamic (RSD) layer for attention modulation. This architectural design enables the model to efficiently process and interpret a range of cross-modal inputs, yielding a comprehensive understanding of the correlation between verbal commands and corresponding visual scenes. Empirical evaluations on the Talk2Car dataset, a real-world benchmark, demonstrate that CAVG establishes new standards in prediction accuracy and operational efficiency. Notably, the model exhibits exceptional performance even with limited training data, ranging from 50\% to 75\% of the full dataset. This feature highlights its effectiveness and potential for deployment in practical AV applications. Moreover, CAVG has shown remarkable robustness and adaptability in challenging scenarios, including long-text command interpretation, low-light conditions, ambiguous command contexts, inclement weather conditions, and densely populated urban environments.
The code for the proposed model is available at our \hypersetup{hidelinks}\href{https://github.com/Petrichor625/Talk2car_CAVG.git}{\color{purple}{Github}}.
\end{abstract}


\section{Introduction}
While autonomous vehicles (AVs) herald a transformative era in urban mobility and have the potential to significantly reduce road accidents, their widespread acceptance remains hampered by public skepticism \cite{li2023context}. The root of this hesitancy often lies in concerns about trust and the perceived loss of control. These issues are exacerbated when AVs display indecisiveness or ambiguity in complex driving scenarios \cite{li2022lane}. To address this trust gap, our research introduces a pioneering framework that augments human-AV interaction by effectively grounding natural language commands within visual contexts. Envision a scenario where a commander instructs an AV with a simple command like "Park in a shaded area on a sunny day." Such interactions not only empower the commander but also infuse the vehicle's decision-making process with essential context, especially in ambiguous or complex driving situations. Recent studies further highlight the significant enhancement in commander experience and acceptance when natural language commands facilitate clear communication of intentions to AVs \cite{deruyttere2021giving, othman2021public, bonnefon2016social, dong2023did}.

However, decoding commander intent poses a substantial challenge for AVs, primarily due to the intricacies of interactions between commanders and AVs. To effectively comprehend and act upon these commands, AVs must adeptly capture the complex semantic relationship between natural language and the dynamic visual environment. This task is further complicated by the multifaceted nature of real-world traffic scenes, characterized by ever-changing visual elements such as traffic agents, lighting conditions, road markings, and fluctuating weather. These elements, coupled with the inherent ambiguity of natural language commands, can lead to potentially hazardous misunderstandings between commanders and AVs, resulting in unsafe driving maneuvers.

\begin{figure*}[t]
 \centering \includegraphics[width=0.8\linewidth]{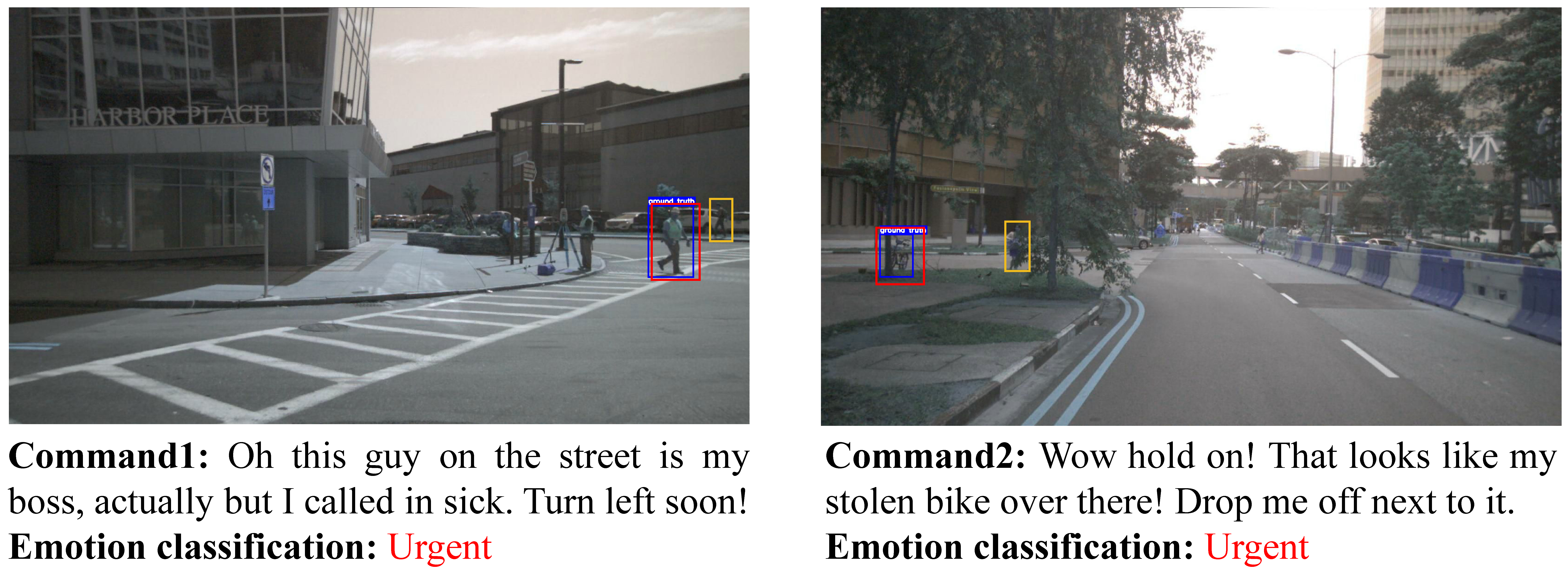} 
 \caption{Illustration of Regions Identified by an AV based on a Raw Image and a Natural Language Command. The blue bounding box represents the ground truth. The red and yellow bounding boxes correspond to the prediction results from CAVG with emotion categorization and without emotion categorization, respectively.}
 \label{toutu} 
\end{figure*}

Addressing these challenges necessitates the development of robust models capable of interpreting and grounding natural language commands amidst this complex and continuously evolving visual landscape. Moreover, considering the emotional and psychological dimensions of human-AV interactions is critical \cite{tang2015learning}. Natural language commands inherently carry a spectrum of emotional overtones—from informative and commanding to urgent. Neglecting the emotional aspect and its fluctuating responses in different traffic contexts could hinder the ability to discern subtle variations in commands. These emotional variables, intertwined with language use, significantly influence the underlying intent and nuances conveyed in commands. Consequently, adopting a holistic approach that encompasses the emotional dimension of human-AV interaction is imperative.

Despite considerable progress in this field, with various methodologies and frameworks proposed \cite{mittal2020attngrounder, deruyttere2020giving, rufus2020cosine}, existing solutions typically rely on a segmented process of object identification followed by linguistic alignment. These methods focus primarily on the bounding boxes of detected objects, often overlooking essential real-world driving contexts. Furthermore, traditional approaches usually concentrate on surface-level features, neglecting underlying elements that could enrich the interpretation and execution of commands. In contrast, recent advancements \cite{bhattacharyya2022aligning,dong2021multi,wen2021cookie} have adopted integrated frameworks that merge textual and visual elements early in the process but face limitations in handling complex multi-object scenarios. Even with the introduction of Transformer-based models \cite{ding2021vision, ding2022vlt, dong2022development}, challenges remain in navigating intricate real-world situations. Importantly, most state-of-the-art models \cite{grujicic2022predicting, chan2022grounding, jain2023ground} in language grounding tend to overlook the emotional dynamics inherent in human-AV interactions, a crucial aspect that our research aims to address.

Our model introduces a novel approach that fuses advancements in visual-language tasks and image-text retrieval. Adopting a sophisticated, two-pronged strategy, it capitalizes on the strengths of both single-stream and dual-stream architectures. The first strategy enhances contextual understanding by integrating contextual cues into the CAVG model, elevating its ability to interpret and respond to natural language commands in real-world driving scenarios. The second strategy involves implementing a multimodal prediction framework, allowing the system to anticipate various plausible outcomes, each paired with a calculated likelihood. This approach is particularly beneficial in mitigating risks and uncertainties in complex and ambiguous driving situations. Additionally, CAVG categorizes subtle emotions inherent in human language, shaping its responses to align with the psychological and emotional context presented by the commanders. As illustrated in Fig. \ref{toutu}, CAVG excels at grounding semantically matched objects in dynamic visual traffic scenes in response to natural language commands, adapting its responses to the emotional context of the commander while maintaining contextual accuracy.

In conclusion, our research introduces a novel model that revolutionizes the way commanders interact with AVs through natural language commands. This innovative approach promises to transform the landscape of human-AV interaction significantly, enhancing commander comfort, enriching user experience, and elevating satisfaction levels. Our model's distinct contributions are as follows:
\begin{itemize}
 \item We have developed a unique approach utilizing a pre-trained model to blend semantic information with global image features. This results in a cross-modal vector, which forms the basis of our novel multi-head cross-model attention mechanism. This mechanism adeptly integrates global context and textual information, enriching the understanding of each sub-region in the visual field.
 \item  Our model incorporates an innovative emotion analysis component using Generative Pretrained Transformer 4 (GPT-4). This allows us to capture the nuanced emotional content in natural language commands. This feature makes the interaction between humans and AVs more intuitive and responsive, enabling the vehicle to understand and react to the commander's intent in a more human-centric manner.
 \item The Cross-Modal Encoder in our model facilitates the extraction of interactive features between different input types, such as images and commands. Utilizing a Region-Specific Dynamic (RSD) layer attention mechanism, our model adeptly analyzes and prioritizes multi-modal data, offering a more refined understanding of the environment.
 \item Our empirical evaluation demonstrates that our model surpasses existing state-of-the-art methods in performance, particularly on the challenging Talk2Car dataset. This includes scenarios like low-light conditions, long-text commands, rainy weather, and dense urban environments. Our model achieves this with less training time, superior performance, and enhanced interpretability.
\end{itemize}

This paper is structured in the following manner to facilitate a comprehensive understanding: Section \ref{Related work} offers a detailed review of the existing literature and studies pertinent to our research area. In Section \ref{Proposed Model}, we articulate the research task at hand and present a detailed diagram of our model's architecture. Section \ref{Experiments} elaborates on the methodology of our experimental setup, followed by an in-depth discussion of the results obtained. In Section \ref{User_study}, we investigate the utility and user satisfaction of our proposed model using a questionnaire. Finally, Section \ref{Conclusion} synthesizes the main findings of this study and proposes potential directions for future investigations in this field.

\section{Related Work}\label{Related work}

\subsection{Visual Grounding}
Visual Grounding (VG) is a task that fundamentally involves localizing image regions pertinent to a specific natural language command. The body of research in this domain typically falls into two primary categories based on their methodological frameworks: one-stage and two-stage methods.

\textbf{One-Stage Methods in Visual Grounding:} These methods are characterized by their efficiency, primarily due to a streamlined processing paradigm. Unlike the two-stage models, which initiate with proposing candidate regions followed by object classification refinement, one-stage methods \cite{chen2018real, yang2019fast, liao2020real, yang2020improving, yang2022improving} simultaneously extract image and command features in a unified pass, predicting both object categories and their bounding boxes concurrently. For instance, \cite{chen2018real} introduced a guided attention mechanism to focus on the referent's central region, interpreting referring expressions in a single stage without resorting to region proposals. \cite{liao2020real} reformulated referring expression comprehension as a correlation filtering process for precise object localization. Moreover, \cite{yang2022improving} integrated a visual-linguistic verification module within a unified one-stage framework to effectively identify regions corresponding with descriptions. The elimination of the region proposal phase in single-stage methods facilitates faster inference times but often requires extended training durations and may involve trade-offs in performance. Particularly, these methods may struggle in scenarios with densely packed objects or overlapping regions of interest, posing accuracy challenges and limiting their robust application in AV contexts.

\textbf{Two-Stage Methods in Visual Grounding:} In contrast, two-stage methods are renowned for their meticulous object identification process within images. These methods typically employ pre-trained models as object detectors in the first stage to extract region-specific features \cite{wang2019neighbourhood, tan2019lxmert, wang2018learning, zhuang2018parallel}. For example, MAttNet \cite{yu2018mattnet} and Volta \cite{bugliarello2021multimodal} utilize Faster R-CNN \cite{ren2015faster} and ResNet-101 \cite{he2016deep} as their backbone networks to capture region features. These features form the basis for defining the spatial locations of potential objects of interest, termed region proposals. In the second stage, these proposals are integrated with linguistic features and analyzed by the framework’s decoder, which selects regions that align closely with the linguistic input, ensuring a semantically coherent and contextually appropriate detection result.

While two-stage methods have demonstrated efficacy in VG tasks, they are often heavily reliant on the accuracy of pre-trained object detection models. This reliance can be a drawback, especially when issues in bounding box localization occur or when crucial semantic information outside the proposed regions is overlooked. Our study adopts a two-stage approach but extends its focus beyond traditional bounding boxes. We aim to capture a broader spectrum of contextual information within various traffic scenarios. This approach seeks to leverage the synergies between bounding box localization and contextual cues, enhancing object localization and promoting context-aware recognition in AV applications.

\subsection{Language-to-Image in Autonomous Driving}
The fusion of object detection with semantic information processing has catalyzed significant advancements in AVs, particularly in the realm of language grounding. A seminal work in this area is the Talk2Car dataset introduced by Deruyttere et al. \cite{deruyttere2019talk2car}. This innovative dataset, comprising natural language commands specifically tailored for self-driving cars, serves as a critical benchmark and sheds light on the challenges of object referral in visual grounding (VG) for AVs.

Progressing from the foundations established by the Talk2Car dataset, recent research has further advanced the field. Mittal's AttnGrounder \cite{mittal2020attngrounder}, for instance, is an end-to-end model with a Cross-Modal attention module that correlates words in a query to regions in visual scenes. Rufus et al. \cite{rufus2020cosine} contributed a minimalist yet effective approach to VG in AVs. Further enriching this domain, Deruyttere et al. \cite{deruyttere2020commands, deruyttere2021giving} and Rufus et al. \cite{rufus2021grounding} explored uncertainties in commands and navigable region identification. Chan et al. \cite{chan2022grounding} and Cheng et al. \cite{cheng2023language} have also made significant strides, with Cheng et al. introducing LiDAR Grounding. Dong et al. \cite{dong2023hubo} presented the HuBo-VLM model, showcasing a unified vision-language approach for human-robot interactions.

\subsection{Emotion Classification}

Emotion classification has been a focus in several studies, including those analyzing emotions expressed in specific events \cite{li2023mitigating, li2022force}. Devlin et al.'s BERT \cite{devlin2018bert} integrated transformer encoders for pretraining models on NLP tasks, including emotion classification. Yang et al. \cite{yang2019emotionx} introduced EmotionX-KU, a classifier for detecting emotions in dialogue, as part of the EmotionX 2019 challenge. Sentiment GPT \cite{kheiri2023sentimentgpt} applied GPT models to emotion classification, highlighting their effectiveness in sentiment analysis. In our study, we utilize GPT-4 for analyzing emotional content in commands, drawing on its extensive pre-training to grasp subtle linguistic patterns and contextual nuances.

\subsection{Vision-and-Language Model}

The domain of Vision-and-Language pre-trained models is characterized by two principal architectures: single-stream and dual-stream. Single-stream models like ImageBert \cite{qi2020imagebert} seamlessly integrate visual and linguistic modalities across encoder layers for holistic data processing. In contrast, dual-stream architectures, such as LXMERT \cite{tan2019lxmert} and VilBERT \cite{su2019vl}, begin with intra-modal interactions, gradually merging modalities as processing progresses. LXMERT, for example, employs an object-relationship encoding module to align complex visual patterns with linguistic cues. In our research, we adopt a hybrid approach that harnesses the strengths of both architectures, optimizing the CAVG model for intricate tasks in AVs.

\begin{figure}[t]
 \centering \includegraphics[width=0.95\linewidth]{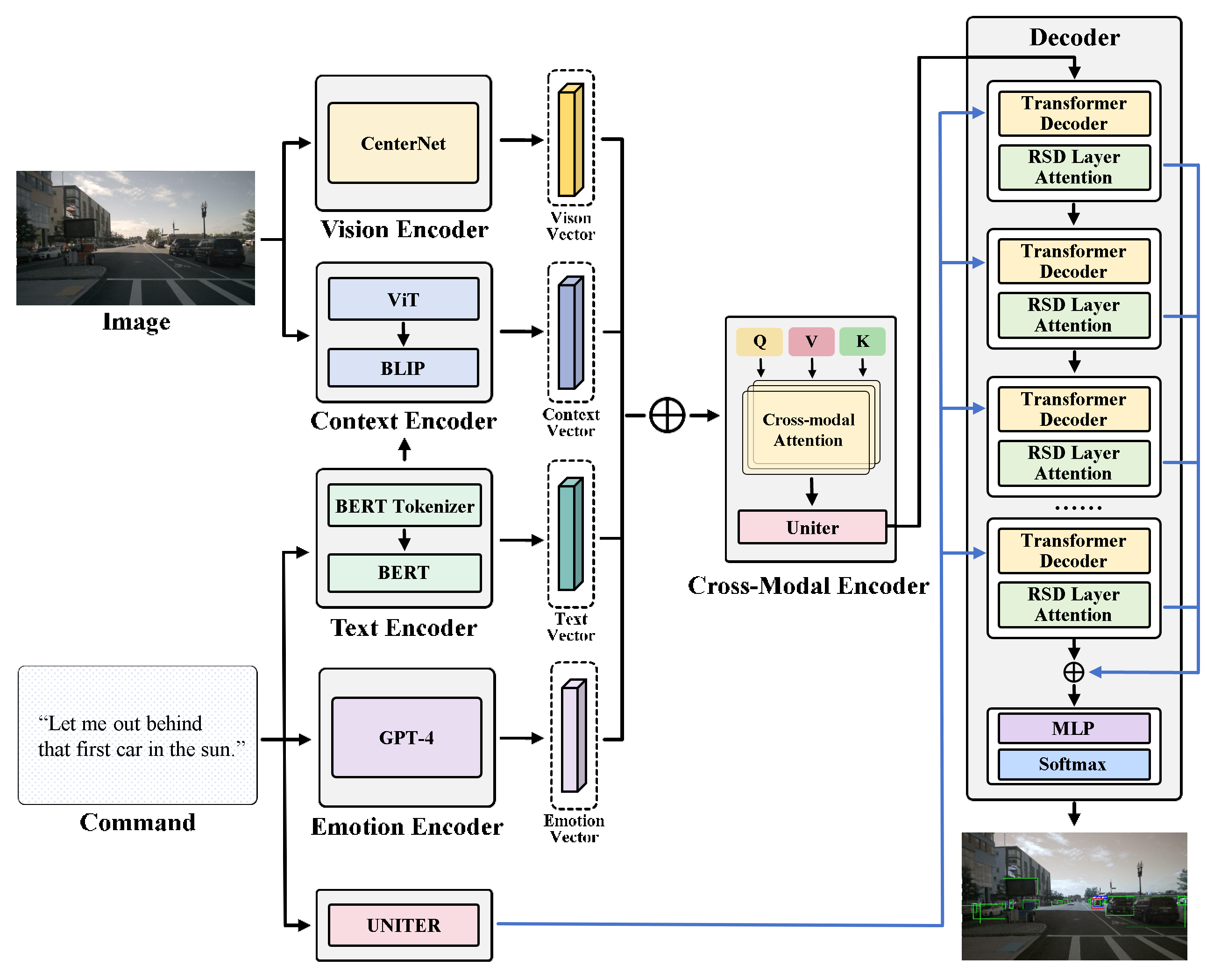} 
 \caption{Schematic of the Model Architecture. The Text Encoder and the Emotion Encoder generate a text vector and an emotion vector, respectively, from the given command, while the Vision Encoder divides the input image into \(N\) RoIs, each represented by a vision vector. These vectors are contextually enriched by a context encoder and then merged by a Cross-Modal Encoder using multi-head cross-modal attention. The multimodal decoder calculates likelihood scores for each region and selects the top-\(k\) regions that best match the semantics of the command. The final prediction is based on this fusion.
 }
 \label{frame_work} 
\end{figure}

\section{Proposed Model}\label{Proposed Model}
\subsection{Overview}
Our aim is to devise a method that can process a frontal-view image of a vehicle, denoted as $I$, in tandem with a natural language command, denoted as $C$. The ultimate objective is to accurately pinpoint the destination region within the image where the vehicle should navigate in order to comply with the given command. 
To formalize this, we cast the task as a mapping problem: identifying the most suitable Regions of Interest (RoIs) in the image $I$ that corresponds to the destination the vehicle must reach based on the natural language command $C$. This mapping is not trivial and involves determining the best-matching bounding boxes within the image that are in line with the semantic intent of the command.

As illustrated in Fig. \ref{frame_work}, our proposed framework utilizes an encoder-decoder architecture that comprises five specialized encoders: the Text Encoder, the Emotion Encoder, the Vision Encoder, the Context Encoder, and the Cross-Modal Encoder. These are integrated with a Multimodal Decoder to perform the task of destination identification.
\subsection{Text Encoder}
The Text Encoder is primarily responsible for processing the natural language command \(C\). 
To achieve this, we use the well-known Bidirectional Transformer (BERT) architecture \cite{devlin2018bert}. BERT has emerged as a seminal advancement in NLP due to its remarkable capabilities in capturing intricate contextual nuances and relationships within textual data. From a technical perspective, BERT uses a multi-layer bidirectional transformer architecture with a deep stack of transformer encoder layers. These layers include self-attention mechanisms that allow the model to capture dependencies between words in both forward and backward directions, ensuring a thorough contextual understanding of the entire input sequence.

Specifically, the Text Encoder uses BERT's robust pre-trained language representations to generate a comprehensive text vector, denoted \(\bm{O}_{\textit{text}}\). The input command \(C\) is tokenized into a sequence using BERT's WordPieces tokenizer and then fed into the BERT model. 
The resulting output is a unified command embedding \(\{w_1, w_2, w_3, \ldots, w_n\}\) that serves as \(\bm{O}_{\textit{text}}\). This embedding encapsulates the textual characteristics of the input command, including its meaning and context. It lays the foundation for subsequent multimodal fusion and contextual analysis, enabling CAVG to seamlessly integrate and process language commands in the realm of autonomous driving. Formally, 
\begin{equation}
\bm{O}_{\textit{text}}= \phi_{\textit{BERT}}\,\left(\phi_{\textit{Tokenizer}}\,(C)\right)= \{w_1, w_2, w_3, \ldots, w_n\}
\end{equation}
where $\phi_{\textit{Tokenizer}}\,(C)$ represents the transformation of the input command $C$ into its corresponding token embeddings using the BERT's WordPiece tokenizer, while the $\phi_{\textit{BERT}}\,(\cdot)$ denotes the BERT architecture. In particular, our Text Encoder is more than just an off-the-shelf implementation of BERT. Instead, we are fine-tuning and adapting BERT to the specific requirements and nuances of language used in the context of autonomous driving. This fine-tuning process adjusts the model to better understand and interpret the unique vocabulary, syntax, and command structures encountered in the context of autonomous driving. 

\subsection{Emotion Encoder}
The Emotion Encoder in our study is crucial for interpreting emotions in commander commands, which is vital for the safe and efficient operation of AVs. This encoder is adept at classifying human emotions, drawing on the advanced capabilities of Large Language Models (LLMs), specifically GPT-4, as referenced by \cite{openai2023gpt4}. GPT-4's pre-training on a broad corpus of text data equips it with a deep understanding of complex linguistic patterns, semantics, and contextual nuances. Its core self-attention mechanism enables nuanced emotion analysis, allowing it to assign different levels of importance to various words in a sentence.

The Emotion Encoder processes commander commands and categorizes them into three distinct emotional states:

\begin{itemize}
 \item \textbf{Urgent:} These commands require immediate action due to their time-sensitive or critical nature. For example, a commander yelling ``Wow hold on! That looks like my stolen bike over there! Drop me off next to it.'', conveys a sense of urgency that requires immediate attention. The Emotion Encoder recognizes this emotional state and interprets this command as requiring immediate attention. 
 Consequently, the destination is inferred as the closest and safest place to pull over, in line with the commander's anxious emotional state.

 \item \textbf{Commanding:} These commands are characterized by their direct and unambiguous nature, specifying what actions the vehicle should take without an inherent time sensitivity. For instance, a command like, ``Make a left turn at the next intersection.'', is clear and commanding. While the command itself is straightforward, our Emotion Encoder detects the absence of urgency in the commander's tone. It categorizes this as a ``commanding'' instruction. In response, the model processes the command accordingly, planning a left turn at the next intersection without the need for immediate action.
 
 \item \textbf{Informative:} Neutral instructions that convey useful information without a strong emotional tone. Informative commands are primarily characterized by their objective nature, focusing on providing factual details about a particular subject or object. These commands tend to be descriptive, where the content predominantly revolves around elucidating specific attributes or features of an entity without conveying a pressing need for action.
\end{itemize}

The Emotion Encoder utilizes the GPT-4 API for emotion analysis. The API setup involves secure communication with OpenAI's servers, and each commander command is preprocessed to align with GPT-4’s input requirements. The command is sent to GPT-4, which then analyzes the text and returns a response indicating the emotional tone. This response is post-processed to categorize the command into one of the predefined categories.

The output of the Emotion Encoder $\bm{O}_{\textit{emo}}$ is the predicted emotion category $e$ for the input command \(C\), which can be defined as follows:
\begin{equation}
\bm{O}_{\textit{emo}}= \phi_{\textit{GPT}}\,(C) = \{e\}
\end{equation}
where the $\phi_{\textit{GPT}}\,(\cdot)$ represents the GPT-4 framework. 
Subsequently, the emotion vector $\bm{O}_{\textit{emo}}$ is fused with the text vector $\bm{O}_{\textit{text}}$ to incorporate the emotional context. This integration is devised to indicate CAVG towards providing human-like responses that resonate with both the given command and the emotional disposition of the commander.

\begin{figure*}[t]
 \centering \includegraphics[width=0.95\linewidth]{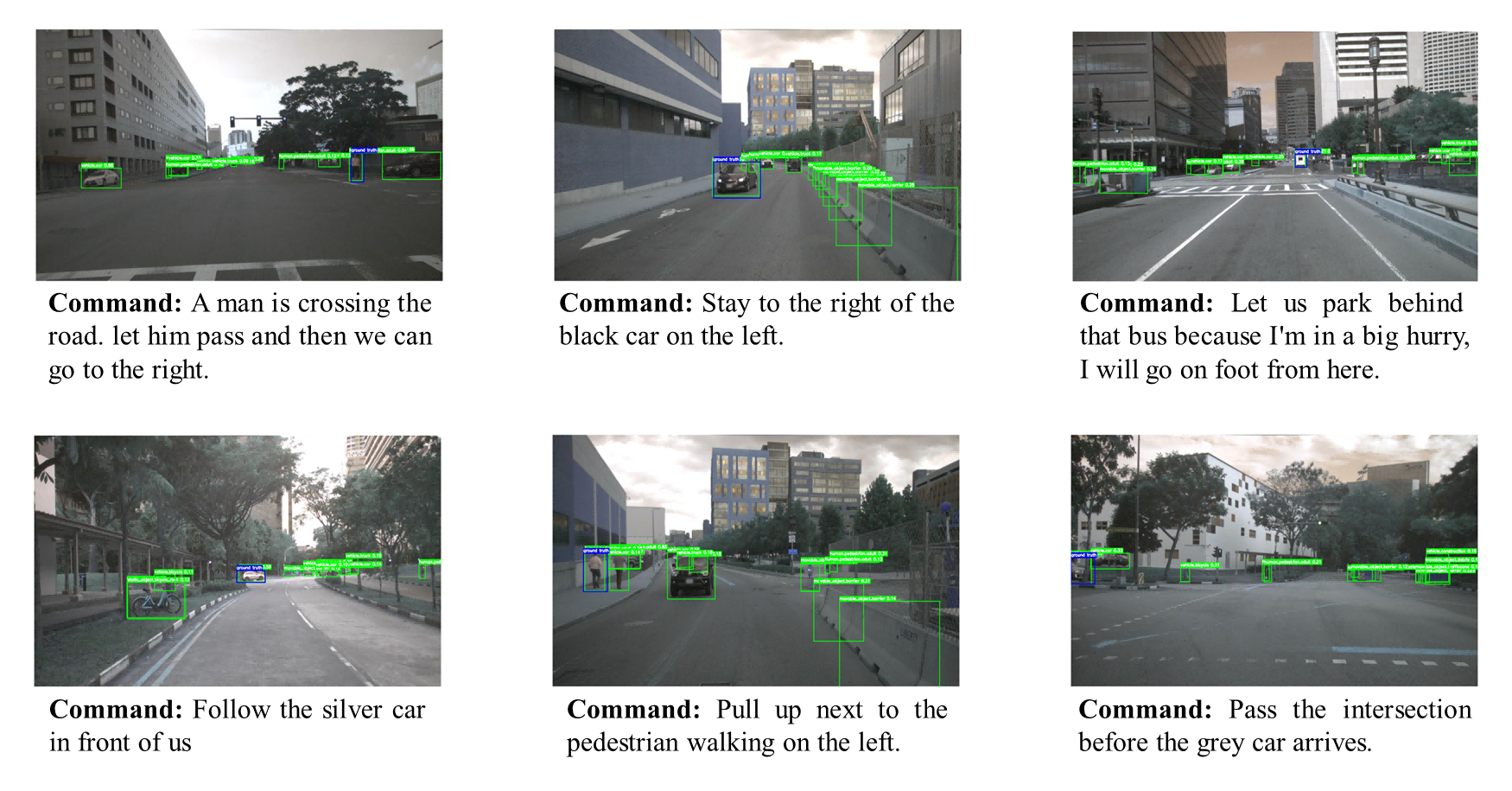}
 \caption{The Output of the Vision Encoder on the Talk2Car Dataset. Each bounding box is labeled with an attribute class followed by an object class. Ground truth bounding boxes are depicted in blue, while the output bounding boxes of CAVG are highlighted in green. A natural language command associated with each visual scenario is also displayed below the image for context.}
 \label{bounding box} 
\end{figure*}

\subsection{Vision Encoder}
The Vision Encoder is architected to extract rich visual information from the given view image \(I\) for subsequent multimodal analysis.
Within this encoder, the CenterNet \cite{dosovitskiy2020image} model is utilized to identify and localize entities and elements within the observed scene in the raw image. Specifically, CenterNet is built on the foundation of two basic components: Fast R-CNN \cite{ren2015faster} and ResNet-101 \cite{he2016deep}. 
Known for its efficiency and accuracy, Fast R-CNN utilizes a region proposal network (RPN) to generate region proposals. These proposals are then subjected to a two-stage process involving a classifier and a regressor. The classifier is responsible for object detection, while the regressor refines the bounding box coordinates. As shown in Fig. \ref{bounding box}, through this process, CAVG can effectively identify prominent entities and detect their classes in the input image, such as trees, vehicles, bicycles, and pedestrians. In particular, these identified entities are encapsulated as RoIs, rich in semantic features and spatial correlations among different entities.

Complementing the capabilities of Fast R-CNN, ResNet-101 is introduced into the framework to enhance feature representation, with a keen focus on capturing finer details, textures, and distinguishing attributes of detected objects. Such nuances are essential for understanding commands that may refer to specific visual cues or unique scene features. For example, given a visual scene with the command ``Could you please park before the red car?'', the ResNet-101 is adept at capturing subtle color variations and textures within the RoIs and embedding them into high-dimensional visual representations. This empowers CAVG to pinpoint the relevant vehicles and greatly helps to provide accurate, context-aware responses.

Overall, this synergistic integration within the CenterNet model enables the encoder to produce the high-dimensional visual representations, encapsulated in a vision vector, labeled \(\bm{O}_{\textit{vision}}\), which can be given as:
\begin{equation}
\bm{O}_{\textit{vision}}= \phi_{\textit{CenterNet}}\,(C) = \{b_1, b_2, b_3, \ldots, b_n\}
\end{equation}
Here, $\phi_{\textit{CenterNet}}\,(\cdot)$ represents the CenterNet model and \(\{b_1, b_2, b_3,\ldots,b_n\}\) is the processed RoIs in the vision vector.

\begin{figure*}[t]
 \centering \includegraphics[width=0.9\linewidth]{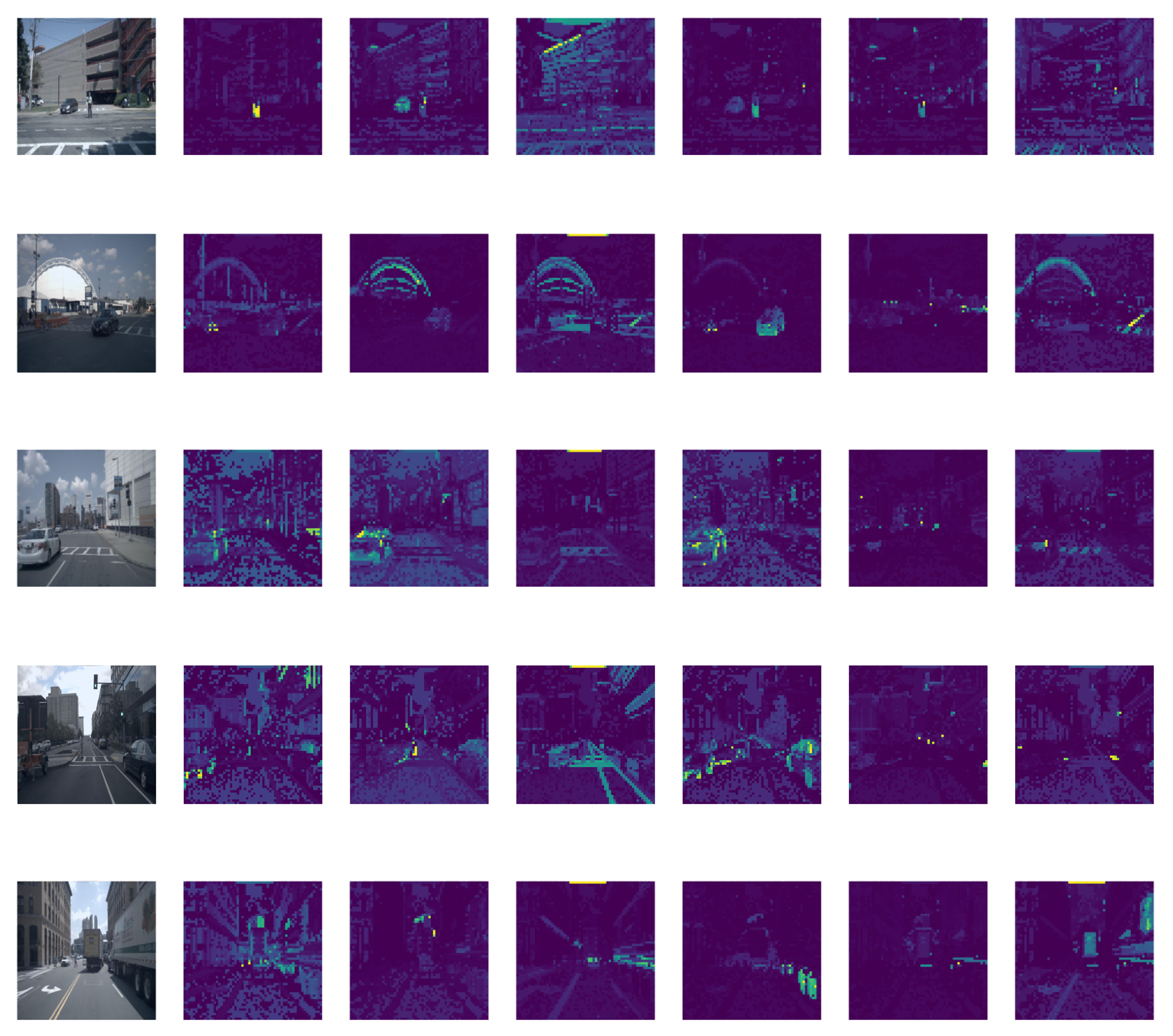} 
 \caption{Raw Images and Examples Output from Context Encoder. CAVG automatically learns to perform object segmentation without a complex self-supervised training recipe or any fine-tuning with segmentation-related annotations. For each image set, we visualize the raw image on the left and the self-attention maps of the image in different layers of the encoder on the right.
 }
 \label{vit_1} 
\end{figure*}

\subsection{Context Encoder}
Our model confronts a more intricate challenge than traditional Referring Image Segmentation tasks due to its multimodal nature and the unstructured nature of its ground truth. In the ``talk-to-car" scenario, the fusion of unrestricted linguistic expressions with diverse visual scenes poses several challenges. These include co-reference resolution, precise identification of named entities, understanding complex object relationships, accurate association of attributes with corresponding visual elements, and pinpointing the object of interest within a command.

Diverging from existing methods that primarily focus RoIs \cite{mittal2020attngrounder,hu2016natural,deruyttere2020giving}, our Context Encoder adopts a more holistic approach. It is designed not only to identify key focal points within the input image but also to discern the broader contextual relationships within the entire visual scene, going beyond the limitations of bounding boxes. To achieve this, we utilize the Vision Transformer (ViT) and the state-of-the-art Vision-and-Language (V\&L) pre-training model, and LLM Bootstrapping Language-Image Pre-training Model (BLIP) \cite{li2022blip}. ViT, in particular, revolutionizes the comprehension capabilities of the Context-Aware Visual Grounding (CAVG) model with its unique ability to consider the entire visual scene as a unified entity. Unlike conventional object-centric methods reliant on bounding boxes, ViT offers a comprehensive perspective of the visual scene. As depicted in Fig. \ref{vit_1}, ViT captures essential contextual details—such as lane markings, pedestrian paths, traffic signs, and other nuances—that are often missed by rigid bounding box constraints. This expanded scope enables our model to assimilate a wider range of visual information and contextual cues, ensuring a thorough semantic interpretation.

In concert with ViT, BLIP acts as a bridge between textual and visual information, enabling CAVG to extract cross-modal features and make meaningful connections between the words in the command and the semantic elements present in the image. This fusion of linguistic context with rich visual cues facilitates a more holistic and nuanced understanding of the scene, enhancing our model's ability to accurately ground textual commands in the visual scenes. The Context Encoder ingests the context vector $\bm{O}_{\textit{context}}$ from text vector $\bm{O}_{\textit{text}}$ and the input image $I$, which can be represented as follows:
\begin{equation}
\bm{O}_{\textit{context}}= \phi_{\textit{BLIP}}\,\left(\phi_{\textit{ViT}}\,(I)\|\bm{O}_{\textit{text}}\right)
\end{equation}
where $\phi_{\textit{BLIP}}$, and $\phi_{\textit{ViT}}$ are the BLIP and ViT frameworks, respectively. In addition, $\|$ denotes matrix concatenation and the context vector $\bm{O}_{\textit{context}}$ not only augments the model with additional visual cues but also broadens its understanding of the scene. This, in turn, enhances the model's ability to make more accurate predictions.

\subsection{Cross-Modal Encoder}
This encoder synergizes the output of the previous encoders through a multi-head cross-modal attention mechanism. Unlike conventional methods like MaskFormer \cite{cheng2021per}, DETR \cite{carion2020end}, and MDETR \cite{kamath2021mdetr}, which are computationally expensive, our approach uses a ``hindsight fusion" strategy. This enables more efficient processing by harmoniously merging visual and text vectors before they are inputted into the Multimodal Decoder.
\begin{figure}[hbtp]
 \centering \includegraphics[width=0.4\linewidth]{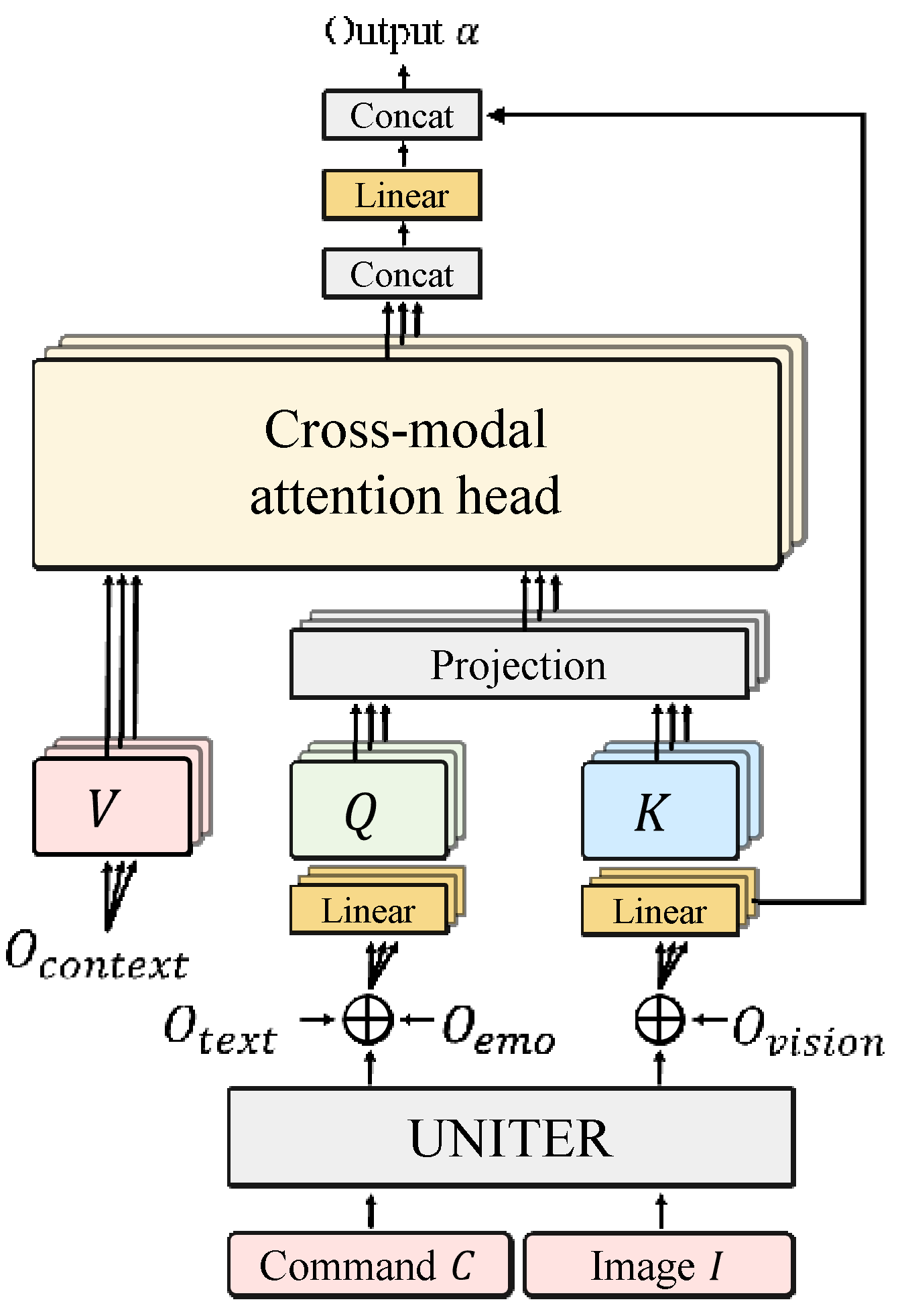} 
 \caption{Overall Framework of Multi-head Cross-modal Self-attention Mechanism.}
 \label{attention} 
\end{figure}
As depicted in Fig. \ref{attention}, the encoder projects the text vector $\bm{O}_{\textit{text}}$ and the emotion vector $\bm{O}_{\textit{emo}}$ into query representation \(Q\) using Multilayer Perceptrons (MLP). Similarly, the vision $\bm{O}_{\textit{vision}}$ and context $\bm{O}_{\textit{context}}$ vectors are converted into the key \(K\) and standardized value \(V\) representations, respectively. To further enhance the text and vision vectors, we append extra tokens $L_{q}$ and $L_{k}$ to the end of the input text vector $\bm{O}_{\textit{text}}$ and vision vector $\bm{O}_{\textit{vision}}$, respectively. Then, these representations are utilized to compute a comprehensive fused vector \(\alpha\), as outlined in the equations. 
In addition, it is worth noting that our cross-modal attention mechanism draws inspiration from the original Transformer \cite{vaswani2017attention} architecture and can also be extended to incorporate multiple attention heads, with a residual connection applied after these attention heads \cite{hao2019sequence}. Formally,
\begin{equation}
\left\{\begin{array}{l}
{Q}_i={W}^{Q} (\phi_{\textit{Linear}}(\bm{O}_{\textit{vision}}+L_{q}))\\
{K}_{i}={W}^{K} (\phi_{\textit{Linear}}((\bm{O}_{\textit{emo}} \|\bm{O}_{\textit{text}})+L_{k})) \\
{V}_{i}={W}^{V} \bm{O}_{\textit{context}}
\end{array}\right.
\end{equation}
where ${W}^{Q}, {W}^{K}, {W}^{V}$ denote the learnable weights, and $\phi_{\textit{Linear}}$ represents the fully connected layer. These tokens $L_{q}$ and $L_{k}$ are generated as position embeddings from the given command $C$ and the raw image $I$ by leveraging the single-stream architecture UNITER (Universal Image-TExt Representation) \cite{chen2020uniter}, which can be expressed as follows:
\begin{equation}
\left\{\begin{array}{l}
L_{q}=\phi_{\textit{UNITER}}\,\left({C}\right)\\
L_{k}=\phi_{\textit{UNITER}}\,\left({I}\right)
\end{array}\right.
\end{equation}
where $\phi_{\textit{UNITER}}$ denotes the UNITER architecture. Moreover, the $i$th self-attention head $\textit{head}_i$ and the output of the multi-head cross-modal self-attention mechanism ${\alpha}$ can be defined mathematically as:
\begin{flalign}
{\alpha} &= \sum_{i=1}^{h} {\textit{head}_i} + \phi_{\text{Linear}}(\bm{O}_{\textit{vision}} + L_{q})\\\nonumber&=\sum_{i=1}^{h} { \phi_{\textit {softmax}}}\left(\frac{{Q}_i ({K}_i)^{\top}}{\sqrt{d_k}}\right) {V}_i + \phi_{\textit{Linear}}(\bm{O}_{\textit{vision}} +L_{q})
\end{flalign}
where $h$ is the total number of attention heads, ${\phi_{\textit {softmax}}}(\cdot)$ is the softmax activation function, and ${d}_k$ represents the dimensionality of the projected key vectors. Furthermore, the output of this attention mechanism ${\alpha}$ is then 
embedded by the UNITER to produce the fused cross-modal vector ${\bar\alpha}$, which can be given as follows:
\begin{equation}
{\bar\alpha}= \phi_{\textit{UNITER}}\,\left({\alpha}\right)
\end{equation}

Our methodology prominently features the UNITER architecture, renowned for its exceptional capacity to align and integrate visual and textual data into a unified representation. Built upon the transformer model, UNITER processes images and text in parallel, transforming images into patch embeddings analogous to text tokens. This method facilitates a seamless integration of linguistic and visual modalities, a crucial aspect for interpreting complex multimodal data in our study. Pretrained on a broad spectrum of V\&L tasks, UNITER can discern a wide range of visual and textual patterns. This extensive pre-training is especially beneficial for our application, where interpreting the intricate interplay between natural language commands and corresponding visual cues is critical. UNITER's ability to create a joint representation of text and images is key to effectively grounding language commands within the specific visual contexts encountered by AVs.

\subsection{Multimodal Decoder}
The final stage of our architecture is the Multimodal Decoder, which plays a pivotal role in synthesizing and interpreting the multimodal data. This decoder integrates the high-level vector ${L_{q}}$, embedded by the UNITER, with the fused cross-modal vector ${\bar\alpha}$. Its primary function is to discern and rank the top-\(k\) RoIs in the image, thereby identifying the most relevant target region for the given command. Comprising \(m\) layers, the Multimodal Decoder is uniquely structured. Each layer is an amalgamation of a Transformer Decoder \cite{vaswani2017attention} and a Region-Specific Dynamic (RSD) Layer Attention Mechanism. The RSD mechanism, inspired by Chan et al. \cite{chan2022grounding}, represents an advanced evolution in attention mechanisms, designed specifically to address the multimodal complexity of our task.

Within each layer, the Transformer Decoder generates a high-dimensional hidden state from its input. This state is then processed by the RSD Layer, a key component that enhances the model's interpretative capabilities. The RSD mechanism dynamically assigns attention weights to the representation vectors of a region across all encoder layers, rather than relying solely on the top layer representation. This approach allows for a more nuanced aggregation of features, ranging from concrete to abstract, across different layers of the encoder. For each region proposal \(v_i\), the RSD Layer evaluates its representation \(h_{l}^{v_i}\) at each encoder layer \(l\), computing relevance scores \(\alpha_{il}\). These scores are used to aggregate the representation vectors across layers, enabling the model to consider a wide array of features for each region and determine its relevance to the command. The hidden states from all layers, each weighted by their computed relevance, are fused into a singular representation. This combined output is then passed through a Multilayer Perceptron (MLP) followed by a softmax activation function, resulting in multiple hypothesis outputs with associated credibility scores. The bounding box with the highest credibility score is selected as the final output. To further enhance feature retention and extraction, skip connections akin to those in ResNet \cite{he2016deep} are employed. These connections feed the high-level vector ${L_{q}}$ into every layer of the Multimodal Decoder. This design allows the model to eliminate redundant features while preserving those essential for accurate interpretation and decision-making, thereby improving the accuracy and reliability of the model's output.

\section{Experiments}\label{Experiments}
\subsection{Dataset}
We assess CAVG using the Talk2Car \cite{deruyttere2019talk2car} dataset, a derivative of the NuScenes \cite{caesar2020nuscenes} dataset, featuring 11,959 natural language commands across 9,217 images of urban environments from Singapore and Boston. The data set is particularly challenging and diverse, captured under various conditions including different times of the day and weather scenarios. It poses intricate tasks like object disambiguation and understanding elaborate text queries, with commands averaging 11 words in length, fitting CAVG's data requirements. Examples include complex instructions like ``This car just passed a crosswalk, slow down to check for pedestrians!''. We utilize it as a benchmark to assess CAVG's accuracy in this study.
\begin{figure*}[htbp]
 \centering \includegraphics[width=\linewidth]{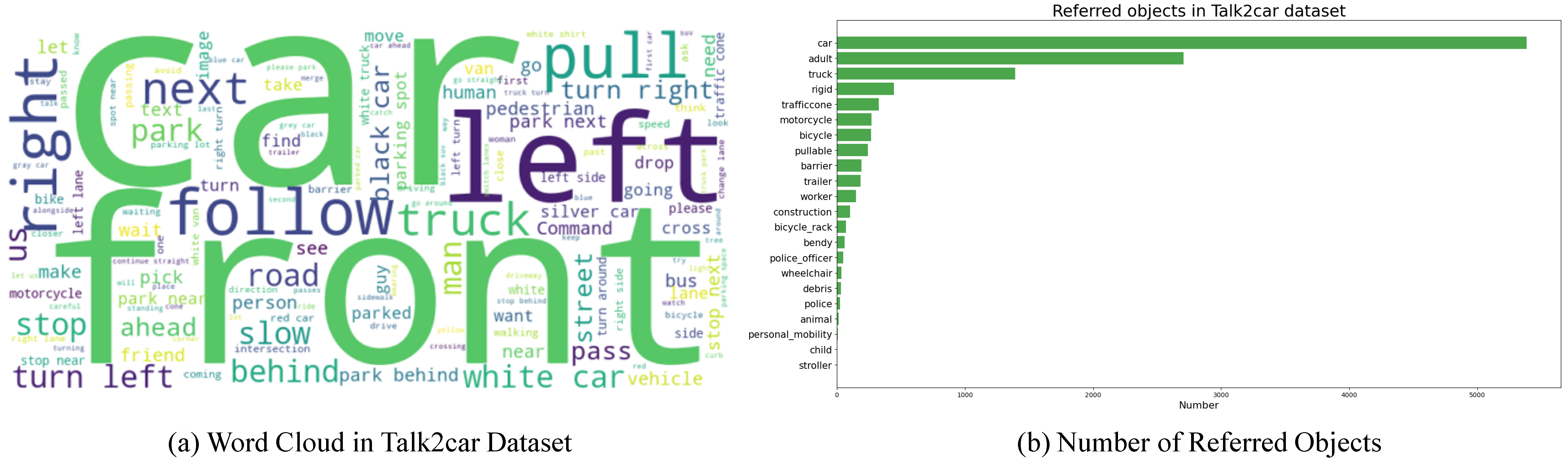} 
 \caption{Statistics of Talk2car Datasets on (a) Word Cloud and (b) Distribution of Referred Object Number.}
 \label{Statistics} 
\end{figure*}
\subsection{Evaluation Metric}
In line with the C4AV challenge, we use the ${IoU}_{0.5}$ score as the evaluation metric. 
It is alternatively referred to as the ${AP}_{50}$ score, which is defined as the percentage of Intersection over Union (IoU) score values that exceed the 0.5 thresholds with the ground truth bounding boxes. This evaluation metric quantifies is widely used in challenges and benchmarks such as the PASCAL \cite{everingham2012pascal}, VOC \cite{everingham2010pascal}, and COCO \cite{lin2014microsoft} datasets, provides a rigorous measure of the accuracy of our predictions, and is consistent with established practices in computer vision and object recognition tasks. Let \( A \) and \( B \) denote the predicted bounding box and the ground truth bounding box, respectively. In this study, each bounding box is represented by the coordinates of the positions of its upper-left \((x_1, y_1)\) and lower-right \((x_2, y_2)\) corners. The following equation elaborates on the computation of IoU between the predicted bounding box \( A \) and the ground truth bounding box \( B \):
\begin{equation}
\text{IoU}(A, B) = \frac{\text{Area of Overlap}\, (A, B)}{\text{Area of Union}\, (A, B)}
\end{equation}

\subsection{Training and Implementation Details}
Our model is trained to converge on an NVIDIA A100 40GB GPU for approximately an hour. It is trained for 6 epochs using the AdamW optimizer, with a batch size of 16 and an initial learning rate set at $10^{-4}$. In addition, the loss function for CAVG is the binary cross-entropy (BCE) loss, with scheduling regulated by the Cosine Annealing Warm Restarts \cite{loshchilov2016sgdr} strategy. Further specifics regarding the implementation and essential parameter settings of CAVG are provided as follows:

\textbf{Text Encoder.} In the text encoder, we use a pre-trained BERT model to generate text embeddings, enforcing a maximum sentence length of 60 tokens. We configure the BERT encoder with 16 hidden layers and an embedding layer with a vocabulary size of 30,524. The epsilon value for the layer normalization is set to 1e-12. The hidden size within the text encoder is set to $d=768$, resulting in a text vector labeled $\bm{O}_{\textit{text}}$ with a dimension of $d$.

\textbf{Emotion Encoder.} 
In this encoder, GPT-4 is used as an emotion classifier to distinguish emotion categories and generate an embedding \(\bm{O}_{\textit{emo}}\) with a dimension of 768, ensuring semantic congruence with \(\bm{O}_{\textit{text}}\). We then concatenate \(\bm{O}_{\textit{emo}}\) and \(\bm{O}_{\textit{text}}\), projecting them to a dimension of $d$.

\textbf{Vision Encoder.} We resize the input image from 1600$\times$900 to 800$\times$450 in the Vision Encoder. Then, Fast R-CNN and ResNet-101 are used as the backbone to locate the bounding boxes of all the objects, and the region proposal threshold is set to 36. Within each region proposal, image information is extracted from the image feature \(\bm{O}_{\textit{vision}}\) with the dimension of 36$\times$1024. 

\textbf{Context Encoder.} The Context Encoder utilizes the complete raw image \(I\) in conjunction with the text vector \(\bm{O}_{\textit{text}}\). Within this framework, we choose an encoder between Vision Transformer-Base and Vision Transformer-Large for the vision embedding layer. Specifically, the ViT-B is configured with a depth of 12, a vision width of 768, a patch size of 16, a ratio of MLP hidden dimension to embedding dimension of 4, and a multi-head attention mechanism with 12 heads. The resulting encoder output is the context embedding \(\bm{O}_{\textit{context}}\) with a dimension of $d$.

\textbf{Cross-Modal Encoder.}
The vectors of text $\bm{O}_{\textit{text}}$, emotion $\bm{O}_{\textit{emotion}}$, vision $\bm{O}_{\textit{vision}}$, and context $\bm{O}_{\textit{context}}$ vectors are fed into this encoder to be converted by the attention mechanism into representations of key, query and values, each with a dimension of 1024. The encoder is configured with 16 heads to optimize the attention mechanism. The UNITER then embeds both \({\alpha}\) and the text feature, using a hidden size of 1024 and producing an output of dimension $d$.

\textbf{Multimodal Decoder.} The Multimodal Decoder is designed with 12 attention layers. Within each layer, we utilize a multi-head attention mechanism configured with 12 heads, and both the vision feature and the text image have a hidden size set to $d$. Additionally, every attention layer incorporates a linear forward progression followed by batch normalization and dropout layers.

\subsection{Experiment Setup}
\begin{figure*}[htbp]
 \centering \includegraphics[width=0.95\linewidth]{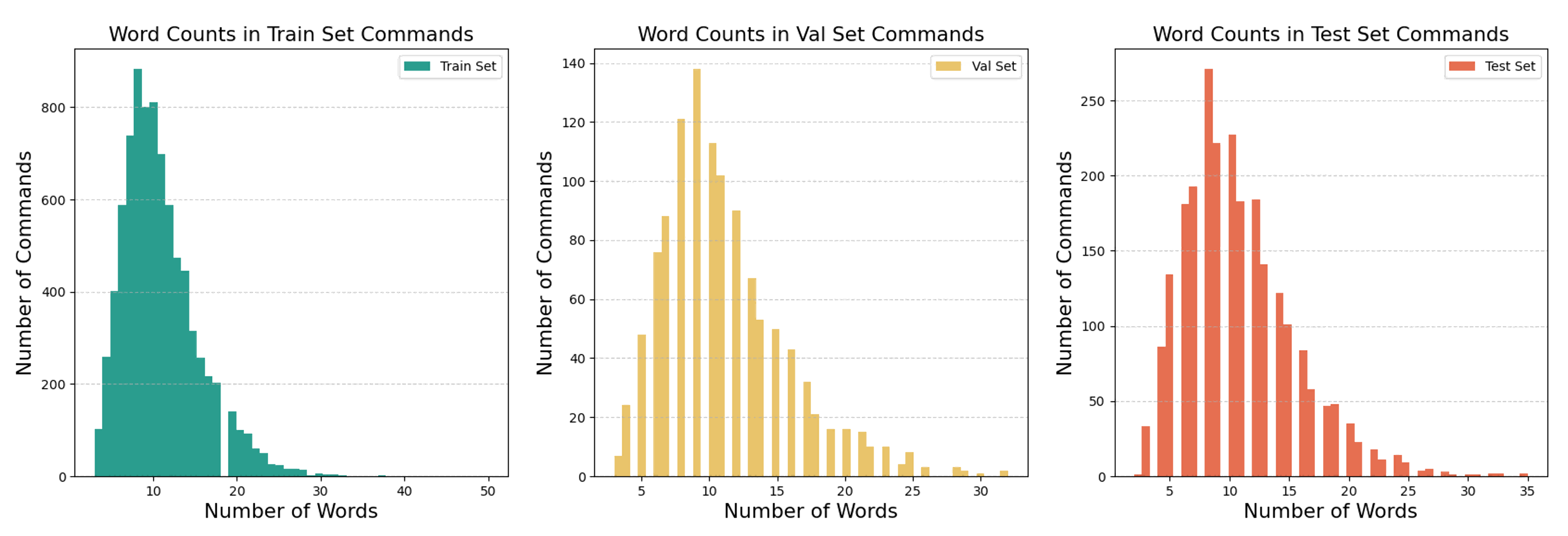} 
 \caption{Word Count Distribution of Commands in the Training, Validation, and Test Set.}
 \label{word_count} 
\end{figure*}

 Our experimental protocol aligns with the Commands For Autonomous Vehicles (C4AV) Challenge protocol. We split the Talk2car dataset into 8,349 training, 1,163 validation, and 2,447 test samples, which constitute 69.8\%, 9.7\%, and 20.5\% of the total dataset. We refer to the complete testset as the \textit{full} testset. With this configuration in mind, we compute the distribution of the word counts for the natural language command on the Talk2Car dataset. The statistical results, shown in Fig. \ref{word_count}, indicate that the training, validation, and testsets exhibit similar data distribution profiles for input commands.

To rigorously evaluate the effectiveness of our model in real-world scenarios, we meticulously partitioned the validation set based on the complexity of linguistic commands and the inherent challenges of visual environments. 
It is worth noting that longer commands may introduce irrelevant details or be more complex for AVs to understand. In addition, specific visual scenarios - such as low-light night scenes, congested urban environments with complex object interactions, ambiguous command prompts, and rainy days where visibility is compromised - invariably complicate prediction.
To this end, we prudently select more than 23 words in length and categorize the aforementioned challenging scenarios into specialized subsets, designated as the \textit{Long-text} and \textit{Corner-case} testsets, respectively. This categorization allows for more focused evaluations.

\begin{figure}[htbp]
 \centering
 \includegraphics[width=0.95\linewidth]{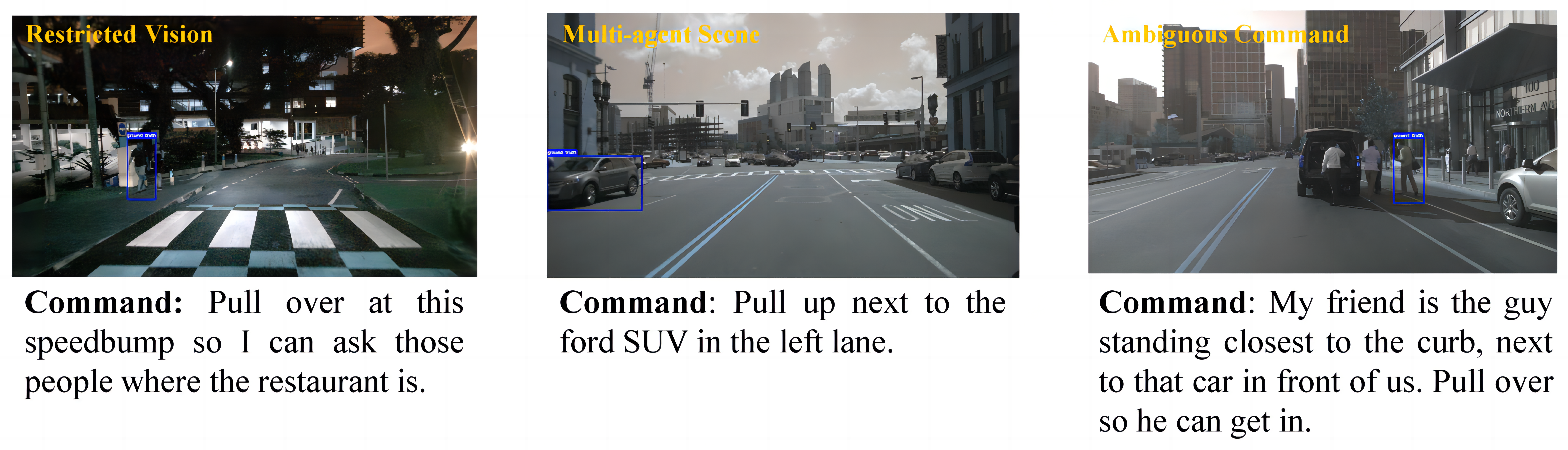}
 \caption{Example for the Subsets of the \textit{Corner-case} Test Set.}
 \label{corner_case}
\end{figure}

In addition, the \textit{Corner-case} testset consists of three main subsets: the scenarios with restricted vision, the scenarios with multi-agent interactions, and the scenarios with ambiguous command prompts, containing 65, 75, and 85 samples, respectively. Figure \ref{corner_case} displays the example scenes for each subset; it is pertinent to mention that a single sample can simultaneously span multiple subsets, provided it is consistent with the definition of multiple scenarios. Correspondingly, we treat the C4AV test set as a ``normal'' subset for comparative analysis. 
Furthermore, for the \textit{Long-text} testset, we employed a data augmentation strategy to increase the richness of the dataset without deviating from the original semantic intent. Using LLM GPT-3.5, we extended certain commands, resulting in commands ranging from 23 to 50 words. This allows us to further evaluate the model's ability to process extended linguistic input, ensuring a comprehensive assessment of our model's adaptability and robustness.

\begin{table}[htbp]
 \centering
 \caption{Performance Comparison of Our Model with Various SOTA Methods on the Talk2Car Dataset using $IoU_{0.5}$ Metric. Models are categorized into three types: one-stage, two-stage, and others, and evaluated based on architectural backbones: \textit{Vision} for visual feature extraction, \textit{Language} for semantic information extraction, and \textit{Global} for holistic data assimilation. Additional components evaluated include emotion classification (EmoClf.), global image feature extraction (Global Img Repr.), linguistic augmentation (NLP Augm.), and visual augmentation (Vis Augm.). `Yes' indicates the use of a feature, `No' its absence, and `-' represents undisclosed data. This categorization elucidates the underlying components and strategies influencing each model's performance. \textbf{Bold} and \underline{underlined} values represent the best and second-best performance, respectively. For comprehensive results, visit the \href{https://www.aicrowd.com/challenges/eccv-2020-commands-4-autonomous-vehicles/leaderboards}{Official Leaderboard}.}
 \resizebox{\textwidth}{!}{
 \begin{tabular}{c|c|ccc|cc|cc}
 \bottomrule
 \multirow{2}[2]{*}{Model} & \multirow{2}[2]{*}{$IoU_{0.5}$} & \multicolumn{3}{c|}{Backbones} & \multicolumn{2}{c|}{Methods} & \multicolumn{2}{c}{Augmentation} \\
 &  & Vision & Language & Global & EmoClf. & Global Img Repr. & NLP Augm. & Vis Augm. \\
 \hline\hline
 \textbf{One-stage:} &  &  &  &  &  &  &  & \\
 MAC  & 50.5 & ResNet-101 & LSTM  & ResNet+CNN & No & Yes  & No & Yes \\
 AttnGrounder & 63.3 & ResNet-53 & LSTM  & Darknet-53 \cite{redmon2018yolov3} & No & Yes  & No & Yes \\
  \hline\hline
 \textbf{Two-stage:} &  &  &  &  &  &  &  & \\
 OSM & 35.3  & ResNet-18  & GRU  & ResNet-18  & No & Yes  & -  & - \\
 SCRC & 43.8  & ResNet-18  & GRU  & ResNet-18  & No & Yes  & -  & - \\
 Bi-Dir.retr & 44.1 & ResNet-18 & LSTM  & No & No & No & No & No \\
 MSRR (Top-16) & 59.9 & ResNet-101 & LSTM  & No & No & No & No & No \\
 MSRR & 60.1 & ResNet-101 & LSTM & No & No & No & No & No \\
 ASSMR (4th) & 66.4 & ResNet-18 & GRU & ResNet-101 & No & Optional & Yes  & Yes \\
 CMSVG (3rd) & 68.6 & EfficientNet \cite{tan2019efficientnet} & Sent.-Transf \cite{reimers2019sentence} & No & No & No & No & No \\
 CMRT (2nd) & 69.1 & ResNet-152 & Transformer \cite{vaswani2017attention} & ResNet-152 & No & Yes  & - & - \\
 Sentence-BET & 70.1 & ResNet-152 & Sent.-Transf \cite{reimers2019sentence} & ResNet-152 & No & Yes  & - & - \\
 Stacked VL-BERT (1st) & 71.0 & ResNet-101 & VL-BERT & ResNet-101 & No & Yes  & - & - \\ 
 \hline\hline
 \textbf{Others:}&  &  &  &  &  &  &  &  \\
 C4AV-Base & 44.3  & -  & -  & -  & No  & -  & No  & - \\
 A-AAR & 45.1  & VGG-16\cite{simonyan2014very}  & Hier.-LSTM \cite{lu2016hierarchical}  & -  & No  & Yes  & -  & - \\
 MaskObjs (Top-16) & 54.3  & -  & -  & -  & No  & -  & -  & - \\
 VL-BERT (base) & 68.3 & -  & -  & -  & No  & -  & -  & - \\
 \hline\hline
  SOTA (Top 1-6) & {66.4}±5.6  &  & &  &  &  & & \\
 SOTA (Top 7-12) &{50.1}±6.8  & &  & &  &  &  &  \\
 \hline\hline
 CAVG (50\%) & 70.3 & CenterNet & BERT/BLIP & UNITER/BLIP & Yes  & Yes  & Yes  & Yes \\
 CAVG (75\%) & \underline{72.1} & CenterNet & BERT/BLIP & UNITER/BLIP & Yes  & Yes  & Yes  & Yes \\
 CAVG & \textbf{74.6} & CenterNet & BERT/BLIP & UNITER/BLIP & Yes  & Yes  & Yes  & Yes \\
  \toprule
 \end{tabular}%
 }
 \label{table2}%
\end{table}%

\subsubsection{Performance Benchmarking Against SOTA Models on \textit{full} Testset}
In our research, we rigorously evaluated CAVG's performance against a comprehensive array of state-of-the-art (SOTA) methods. These include well-known benchmarks such as STACK \cite{hu2016natural}, OSC \cite{vandenhende2020baseline}, SCRC \cite{vandenhende2020baseline}, C4AV-Base \cite{hudson2018compositional}, A-ARR \cite{deng2018visual}, MAC \cite{hudson2018compositional}, MaskObjs (Top-16) \cite{deruyttere2020giving}, MSRR \cite{deruyttere2020giving}, AttnGrounder \cite{mittal2020attngrounder}, VL-BERT (Base) \cite{vandenhende2020baseline}, and Sentence-BET \cite{grujicic2022predicting}. Additionally, we included the top four contenders from the prestigious C4AV Challenge \cite{deruyttere2019talk2car}—namely, Stacked VL-BERT (1st) \cite{dai2020commands}, CMRT (2nd) \cite{luo2020c4av}, CMSVG (3rd) \cite{rufus2020cosine}, and ASSMR (4th) \cite{ou2020attention}—to ensure a holistic assessment. The complete performance metrics for this comparative analysis are available in Table \ref{table2}. 

Remarkably, within the \textit{full} testset, our model did not just hold its own—it dominated across the board. Not only did it surpass all the aforementioned SOTA methods, but it also eclipsed the performances of all models from the C4AV Challenge leaderboard. In terms of the \({IoU}_{0.5}\) metric, the superiority of CAVG is clear: it outperformed Stacked VL-BERT (1st) by a margin of 4.8\%, outperformed CMRT (2nd) by 7.4\%, eclipsed CMSVG (3rd) by 7.9\%, and dramatically outperformed ASSMR (4th) by an impressive 11.0\%. This compelling empirical evidence strongly attests to the exceptional efficacy of our proposed model in the task at hand.

\subsubsection{Performance Evaluation of Training on Reduced Datasets}
In order to showcase the scalability and efficiency of CAVG, we conducted an experiment in which the model was trained on different subsets of the training set, specifically 50\% and 75\%. As illustrated in Tables \ref{table2}, under the labels CAVG (50\%) and CAVG (75\%), CAVG exhibits commendable performance. When conditioned on the 75\% subset of the training data, CAVG consistently outperforms all other baseline models, outperforming C4AV SOTA baselines Stacked VL-BERT (1st) and CMRT (2nd) by 5.07\% and 7.96\%, respectively. 
Impressively, an interesting finding is that when the model is trained on only 50\% of the training data, CAVG displays an admirable performance by achieving significantly lower \({IoU}_{0.5}\) values compared to most of the baselines, and outperforms almost all of the top 1-6 SOTA baselines. These results underscore the potential of CAVG to provide competitive performance even in the absence of extensive training data. This property is of paramount importance, as it suggests potential avenues for economizing data acquisition and annotation efforts, particularly relevant for the autonomous vehicle industry, where data acquisition can be both time-consuming and costly. 
\begin{figure*}[t]
 \centering \includegraphics[width=0.9\linewidth]{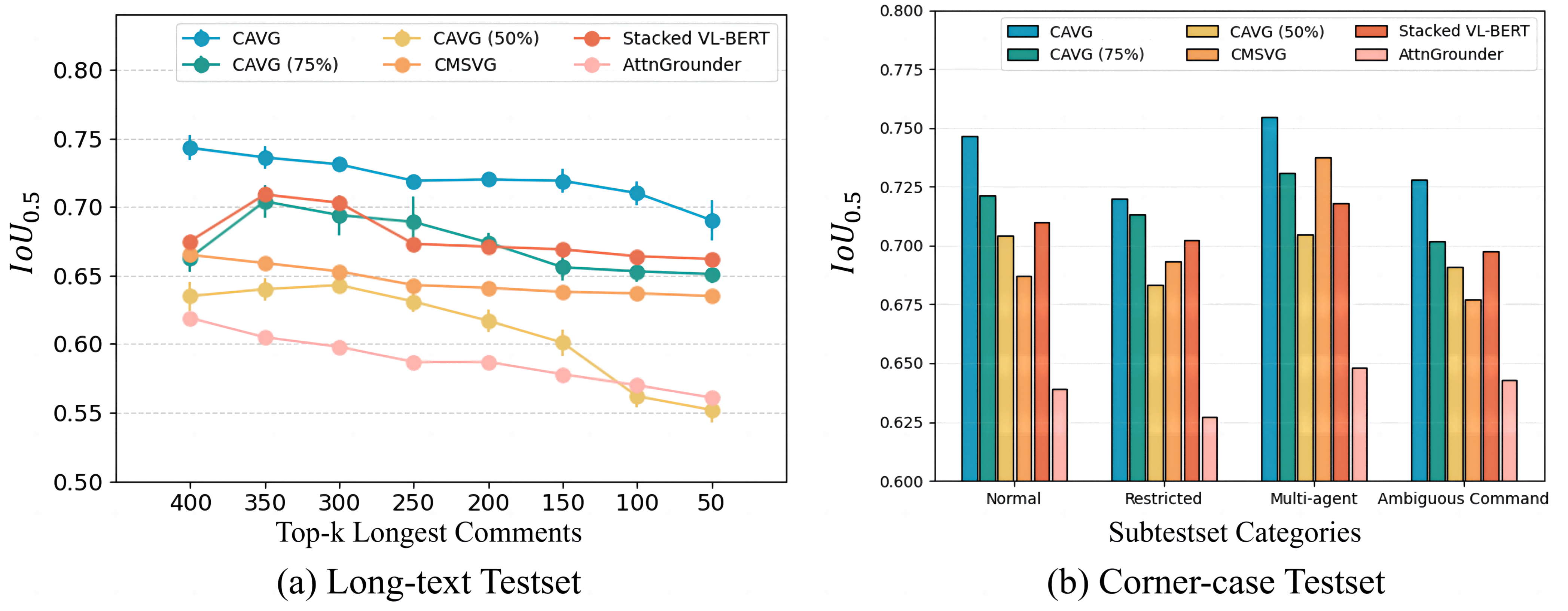} 
 \caption{Evaluation Results in \textit{Long-text} and \textit{Corner-case} Testsets.}
 \label{test_results} 
\end{figure*}

\subsubsection{Performance Evaluation on \textit{Long-text} and \textit{Corner-case} Testsets}
For a holistic insight into the robustness and adaptability of CAVG, we performed further evaluations on the \textit{Long-text} and \textit{Corner-case} testsets, as detailed in Table \ref{long_text} and Table \ref{corner-case}, respectively. To our knowledge, few methods in this area provide the code for these proposed models. Therefore, our comparison focuses on competing open-source models. When confronted with the challenges of the \textit{Long-text} testset, as shown in Fig. \ref{test_results} (a), CAVG consistently surpasses all other baselines, showing an improvement of 4.0\%-10.1\% over the SOTA baselines. Notably, even our model variant trained on only 75\% of the dataset, CAVG (75\%), shows commendable performance, rivaling if not surpassing the current SOTA model, Stacked VL-BERT (1st).  Furthermore, the iteration of CAVG trained with only half the dataset, CAVG (50\%), shows impressive performance, outperforming the majority of the established models. Interestingly, our models - CAVG and CAVG (75\%) - excel in multi-agent scenarios compared to normal ones when provided with sufficient training data. These results highlight their superior ability to understand and interpret commands of considerable length and complexity.

Shifting our gaze to the \textit{Corner-case} testset in Fig. \ref{test_results} (b), the strengths of CAVG become even more apparent in the challenging scenes. 
It exceeds current SOTA baselines, with accuracy gains ranging from 1.9\%-4.9\% on the Talk2car dataset. Moreover, even the CAVG trained on only 75\% of the dataset outperforms other SOTA baselines, while its counterpart, the CAVG trained on only 50\% of the dataset, still manifests commendable performance. This demonstrates the inherent robustness of CAVG, even under challenging conditions, and underscores its promise for use in real-world AV scenarios.

\begin{table}[htbp]
  \centering
  \caption{Performance Comparison of Our Model with Open-source SOTA Methods on the \textit{Long-text} Testset using $IoU_{0.5}$ Metric. \textbf{Bold} and \underline{underlined} values represent the best and second-best performance, respectively.}
    \resizebox{0.7\textwidth}{!}{
    \begin{tabular}{c|c|c|c|c|c|c|c|c}
   \bottomrule
    \multirow{2}[2]{*}{Model} & \multicolumn{8}{c}{Top-k Longest Comments} \\
          & \multicolumn{1}{c}{400} & \multicolumn{1}{c}{350} & \multicolumn{1}{c}{300} & \multicolumn{1}{c}{250} & \multicolumn{1}{c}{200} & \multicolumn{1}{c}{150} & \multicolumn{1}{c}{100} & 50 \\
     \hline
  \hline
    AttnGrounder & 61.9  & 60.5  & 59.8  & 58.7  & 58.6  & 57.8 & 57.0    & 56.1 \\
    CMSVG & 66.5  & 65.9  & 65.3  & 64.3  & 64.1  & 63.8  & 63.7  & 63.5 \\
    Stacked VL-BERT (1st) & \underline{67.5}  & \underline{70.9}  & \underline{70.3}  & 67.3  & 67.1  & \underline{66.9}  & \underline{66.4}  & \underline{66.2} \\
    CAVG (50\%) & 63.5  & 64.0    & 64.3  & 63.1  & 61.7  & 60.1  & 56.2  & 55.2 \\
    CAVG (75\%) & 66.3  & 70.4  & 69.4  & \underline{68.9}  & \underline{67.4}  & 65.6  & 65.3  & 65.1 \\
    \hline
 
    CAVG (100\%) & \textbf{74.1}& \textbf{73.6} & \textbf{73.1} & \textbf{71.9} & \textbf{72.2} & \textbf{71.9} & \textbf{71.0} & \textbf{69.1} \\
    \toprule
    \end{tabular}%
    }
  \label{long_text}%
\end{table}%

\begin{table}[htbp]
  \centering
   \caption{Performance Comparison of Our Model with Open-source SOTA Methods on the \textit{Corner-case} Testset using $IoU_{0.5}$ Metric.  \textbf{Bold} and \underline{underlined} values represent the best and second-best performance, respectively.}
       \resizebox{0.7\textwidth}{!}{
    \begin{tabular}{c|c|c|c|c}
    \bottomrule
    \multirow{2}[2]{*}{Model} & \multicolumn{4}{c}{Subtestset Categories} \\
          & \multicolumn{1}{c}{Normal} & \multicolumn{1}{c}{Restricted} & \multicolumn{1}{c}{Multi-agent} & Ambiguous Command \\
  \hline
  \hline
    AttnGrounder & 63.9  & 62.7  & 64.8  & 64.3 \\
    CMSVG & 68.7  & 69.3  & \underline{73.8}  & 67.8 \\
    Stacked VL-BERT (1st)& 71.1  & 70.2  & 71.8  & 69.8 \\
    CAVG (50\%) & 70.4  & 68.8  & 70.4  & 69.1 \\
    CAVG (75\%) & \underline{72.1}  & \underline{71.3}  & 73.1& \underline{70.2} \\
 \hline
    CAVG (100\%) & \textbf{74.6} & \textbf{72.1} & \textbf{75.2} & \textbf{72.8} \\
    \toprule
    \end{tabular}%
    }
  \label{corner-case}%
\end{table}%

\subsubsection{Comparative Analysis of Model Inference Speed}  
We evaluate the inference speed of CAVG models for different architectures in the Talk2car dataset. Specifically, we test the inference speed of all models on a single NVIDIA GeForce RTX 3090 24GB GPU for a fair comparison. Table \ref{inference} presents a comparative analysis of various models against the CAVG (baseline), which employs the standard R-CNN \cite{girshick2014rich} as a replacement for the Vision Encoder, omitting the BLIP and Context Encoder. These models are defined as follows, with each compared to the full proposed model CAVG:
\begin{itemize}
    \item CAVG: The full model proposed in this paper.
    \item CAVG (ResNet-152): Uses ResNet-152 in the Vision Encoder for visual feature extraction.
    \item CAVG (ResNet-50): Includes ResNet-50 in the Vision Encoder for vision vector extraction.
    \item CAVG (ResNet-18): Includes ResNet-18 in the Vision Encoder for vision vector extraction.
    \item CAVG (BLIP): Applies BLIP only in the Text and Context Encoders for vision and context vector extraction.
    \item CAVG (Small): A scaled-down model featuring a reduced number of attention heads (4 heads) in the Cross-Modal Encoder with 256d hidden states, and fewer attention layers (4 layers) in the RSD.
\end{itemize}
We obtain the average inference time of CAVG (baseline) is about 0.037s per sample. Meanwhile, using the Vision Encoder can speed up the inference speed, showing at least improvements of 20.1\% compared to the baseline. This result demonstrates that incorporating CenterNet into the Vision Encoder significantly increases processing speed, while achieving better performance in terms of prediction accuracy, underscoring its effectiveness in visual grounding contexts for AVs where accuracy, real-time responses are critical. Despite utilizing smaller ResNet variants in the Vision Encoder, the CAVG (ResNet-18) and CAVG (ResNet-50) models maintain strong predictive capabilities, outshining many SOTA models in the \textit{full} test set. They also significantly boost inference speeds by 31. 5\% and 28. 3\%, respectively, compared to CAVG (baseline). Intuitively, the inference time increases with the larger size of the ResNet used in the Vision Encoder. However, the CAVG (ResNet-152), despite its larger size, still provides faster inference capabilities than the CAVG (baseline). It is noted that the CAVG (Small) achieves an optimal balance between computational efficiency and prediction accuracy. This model is 62.9\% faster in inference than the CAVG (baseline), while also improving prediction performance by 2.1\% compared to Stacked VL-BERT (1st). This highlights the potential of the proposed model to achieve high prediction accuracy with significantly reduced computational requirements.
\begin{table}[htbp]
  \centering
  \caption{Inference Time of CAVG Models for Different Architectures and their Predicted Performance in the Talk2car dataset. CAVG (baseline) is selected as the baseline for comparative analysis. These models are evaluated based on architectural backbones: \textit{Vision} for visual feature extraction, \textit{Language} for semantic information extraction, and \textit{Global} for holistic data assimilation. \textbf{Bold} and \underline{underlined} values represent the best and second-best performance, respectively.}
  \resizebox{\textwidth}{!}{
    \begin{tabular}{cc|c|c|c|c|ccc}
     \bottomrule
    \multicolumn{1}{c|}{\multirow{2}[2]{*}{Model}} & \multicolumn{2}{c|}{Accuracy} & \multicolumn{3}{c|}{Speed} & \multicolumn{3}{c}{Backbones} \\
    \multicolumn{1}{c|}{} & \multicolumn{1}{c}{$IoU_{0.5}$} & Accuracy-up (\%) & \multicolumn{1}{c}{Sample Size} & \multicolumn{1}{c}{Inference Time (s)} & Speed-up (\%) & Vision & Language & \multicolumn{1}{c}{Global} \\
    \hline
  \hline
    \multicolumn{1}{c|}{CAVG (baseline)} & 61.4  & -     & 64    & 2.43212 & -     & \multicolumn{1}{c|}{R-CNN} & \multicolumn{1}{c|}{BERT} & \multicolumn{1}{c}{-} \\
    \multicolumn{1}{c|}{CAVG (ResNet-152)} & 74.1  & 20.68\% & 64    & 1.94372 & 20.1\% & \multicolumn{1}{c|}{ResNet-152} & \multicolumn{1}{c|}{BERT/BLIP} & \multicolumn{1}{c}{UNITER/BLIP} \\
    \multicolumn{1}{c|}{CAVG (ResNet-50)} & 72.2  & 17.59\% & 64    & 1.74321 & 28.3\% & \multicolumn{1}{c|}{ResNet-50} & \multicolumn{1}{c|}{BERT/BLIP} & \multicolumn{1}{c}{UNITER/BLIP} \\
    \multicolumn{1}{c|}{CAVG (ResNet-18)} & 68.8  & 12.05\% & 64    & \underline{1.66528} & \underline{31.5\%} & \multicolumn{1}{c|}{ResNet-18} & \multicolumn{1}{c|}{BERT/BLIP} & \multicolumn{1}{c}{UNITER/BLIP} \\
    \multicolumn{1}{c|}{CAVG (BLIP)} & 71.4  & 16.28\% & 64    & 1.77284 & 27.1\% & \multicolumn{1}{c|}{ResNet-101} & \multicolumn{1}{c|}{BLIP} & \multicolumn{1}{c}{BLIP} \\
    \multicolumn{1}{c|}{CAVG (Small)} & \underline{72.5}  & 18.08\%& 64    & \textbf{0.90134} & \textbf{62.9\%} & \multicolumn{1}{c|}{ResNet-18} & \multicolumn{1}{c|}{BERT/BLIP} & UNITER/BLIP \\
    \hline
  \hline
    CAVG  & \textbf{74.6}  & \textbf{21.50\%} & 64    & 1.90446 & 21.7\% & ResNet-101 & BERT/BLIP & UNITER/BLIP \\
    \toprule
    \end{tabular}%
    }
  \label{inference}%
\end{table}%

\subsection{Case Study for RSD Layer Attention Weights}
We visualize the distribution of layer attention weights for 13 layers in a multimodal decoder to further showcase the effectiveness of our proposed RSD layer attention mechanism. We classify the input regions into two distinct groups based on their $IoU$ alignment with the ground-truth region:
\begin{itemize}
 \item $IoU > 0$: encompassing all regions with an $IoU$ exceeding 0, indicating an overlap with the ground-truth region.
 \item $IoU = 0$: Constituting regions exhibiting no overlap, having an $IoU$ of precisely 0.
\end{itemize}
\begin{figure}[htbp]
 \centering \includegraphics[width=0.75\linewidth]{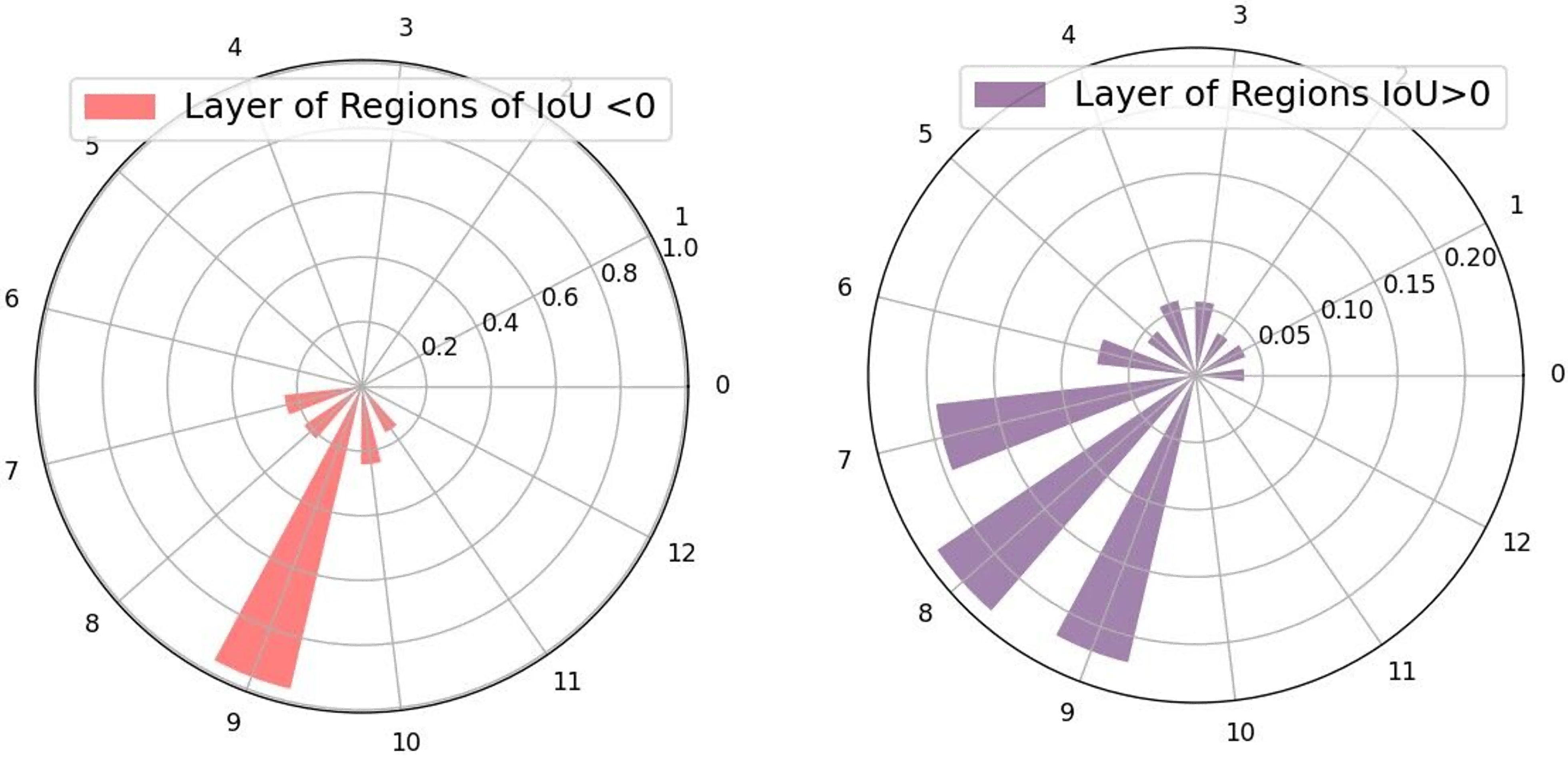} 
 \caption{Attention Weights over all Decoder Layers for RSD Layer Attention. The 0-th layer indicates the embedding layer, while the 1-st denotes the first layer of RSD layer attention, and the 12-th represents the last layer.}
 \label{attention_map} 
\end{figure}
As shown in Fig. \ref{attention_map}, the higher decoder layers (especially layers 7 to 10) are endowed with a larger fraction of attention weights. This observation suggests that vectors have a greater influence on these higher layers, possibly due to increased cross-modal interactions. Interestingly, contrary to our initial expectations, the top layer does not dominate the attention weights. On the contrary, its output appears to be less central. This is a marked departure from traditional techniques that rely predominantly on the topmost layer representation to predict the best-aligned region, potentially sidelining the nuanced cross-modal features inherent in other layers. In light of this, we introduce the RSD layer mechanism, which is designed to dynamically allocate attention weights across all decoder layers, anchored by their hidden states. Subsequently, the model integrates the outputs of different layers and harmonizes them on the basis of these attention weights. This innovative approach equips our model with the ability to capture and exploit richer cross-modal semantic features, which are essential to command and scene understanding.

\subsection{Ablation Study}
\subsubsection{Ablation Study for Core Module}
We conducted a rigorous ablation study to determine the effect of core components in CAVG.
As shown in Table \ref{Table3}, specifically, model H, inclusive of all components, consistently outperforms across the evaluation metric, illustrating the collective value of the components for optimal performance. Model A, a reduced version of Model H, replaces the Vision Encoder with the classical R-CNN, leading to substantial decreases in $IoU_{0.5}$ score by 5.8\%, highlighting the importance of visual features in prediction accuracy. This result also reveals the remarkable superiority of the Vision Encoder structure over the traditional R-CNN model. The Vision Encoder, with its single-stage, anchor-free design, streamlines the object detection process. It locates object centers directly on the feature map and uses regression to determine object dimensions. This approach not only eliminates the need for region proposal and anchor adjustment, simplifying the detection process but also provides a significant increase in detection accuracy. 

Model B, which excludes the Context and Emotion Encoder compared to the full model, shows significant metric reductions, this underscores the essential role of the additional visual cues in improving prediction accuracy. 
Furthermore, Model C, which omits the Text Encoder and Cross-Modal Encoder, exhibits performance drops of 6.6\% in metric, suggesting the importance of processing instructions and multimodal information individually in this task. In addition, Model D, which lacks the visual Text Encoder, suffers notable performance degradation, emphasizing the improved efficacy of the proposed multi-head cross-modal attention mechanism. 
Accordingly, Model E and Model F omit the UNITER embedding process in the Multimodal Decoder and the Emotion Encoder, respectively, resulting in a decrease in the $IoU_{0.5}$ score as well. Finally, Model G, which omits the three core components of CAVG, suffers a noticeable decline in performance compared to Model H. This demonstrates the significant role that the synergistic integration of various modules plays in enhancing the accuracy of our model.
\begin{table}[htbp]
 \centering
 \caption{Component Constituents of Ablation Models.}\label{Table3}
 \resizebox{0.64\linewidth}{!}{
 \begin{tabular}{p{18.625em}cccccccc}
 \toprule
 \multicolumn{1}{c}{\multirow{2}[4]{*}{Components}} & \multicolumn{7}{c}{Ablation Models} \\
\cmidrule{2-9} \multicolumn{1}{c}{} & A & B & C & D & E & F & G &H \\
 \midrule
 \multicolumn{1}{c}{Vision Encoder} & \ding{56} & \ding{52} & \ding{52} & \ding{52} & \ding{52} & \ding{52} & \ding{52} & \ding{52}\\
 \multicolumn{1}{c}{Context Encoder} & \ding{52} & \ding{56} & \ding{52} & \ding{52} & \ding{52} & \ding{52} & \ding{56} & \ding{52}\\
 \multicolumn{1}{c}{Text Encoder} & \ding{52} & \ding{52} & \ding{56} & \ding{52} & \ding{52} & \ding{52} & \ding{52} & \ding{52}\\

 \multicolumn{1}{c}{Cross-Modal Encoder} & \ding{52} & \ding{52} & \ding{56} & \ding{56} & \ding{52} & \ding{52} & \ding{52} & \ding{52}\\
 \multicolumn{1}{c}{UNITER-Embedding} & \ding{52} & \ding{52} & \ding{52} & \ding{52} & \ding{56} & \ding{52} & \ding{56} & \ding{52}\\

 \multicolumn{1}{c}{Emotion Encoder} & \ding{52} & \ding{52} & \ding{52} & \ding{52} & \ding{56} & \ding{56} & \ding{56} & \ding{52}\\
 \midrule
 \multicolumn{1}{c}{$IoU_{0.5}$} & 71.3 & 66.5 & 68.4 & 70.1  & 71.9 & 73.2 & 64.1 & 74.6\\
 \bottomrule
 \end{tabular}%
 }
\end{table}%
\begin{figure}[htbp]
 \centering
 \includegraphics[width=0.9\linewidth]{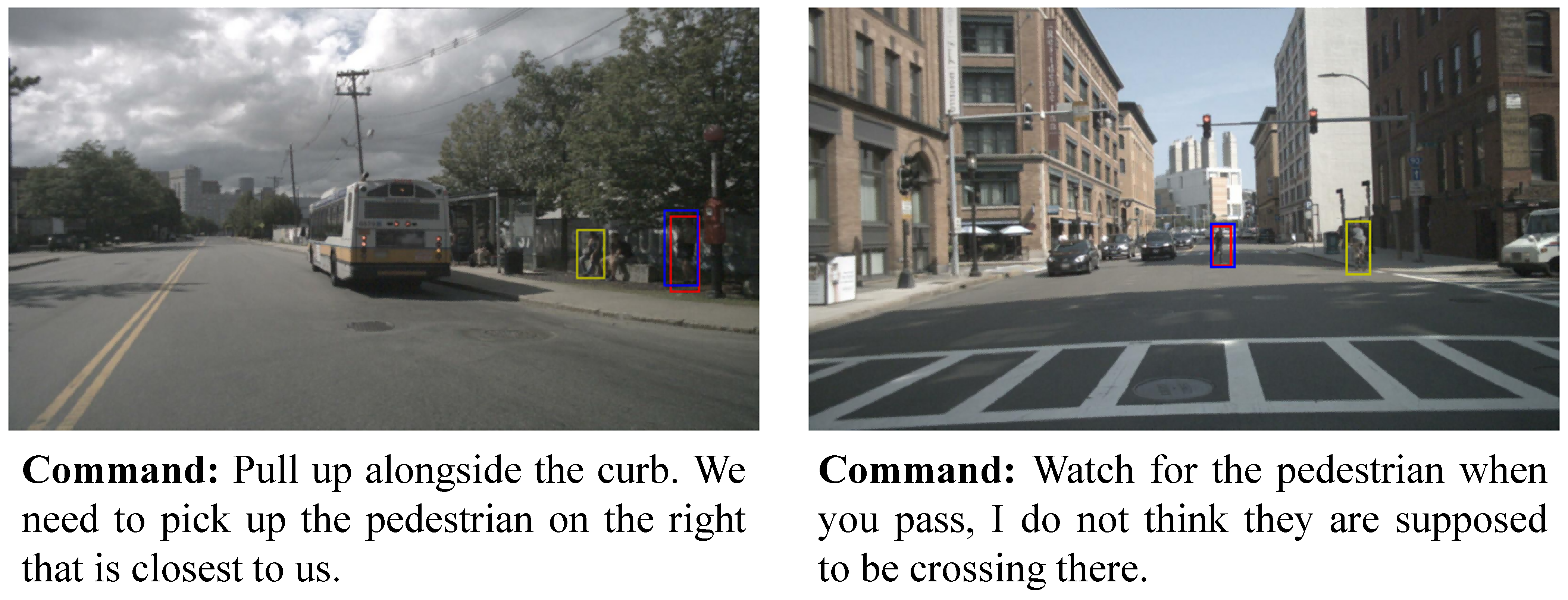}
 \caption{Visualization of Prediction Results for CAVG with and without the Integration of Context Encoder and the Cross-Modal Attention Mechanism. The blue box denotes the ground-truth region of interest. The yellow bounding box represents predictions made by CAVG without the use of a context encoder and multi-head cross-modal attention mechanisms. The red box shows predictions from CAVG when these advanced components are incorporated.}
 \label{failure}
\end{figure}

\begin{figure*}[t]
 \centering \includegraphics[width=0.9\linewidth]{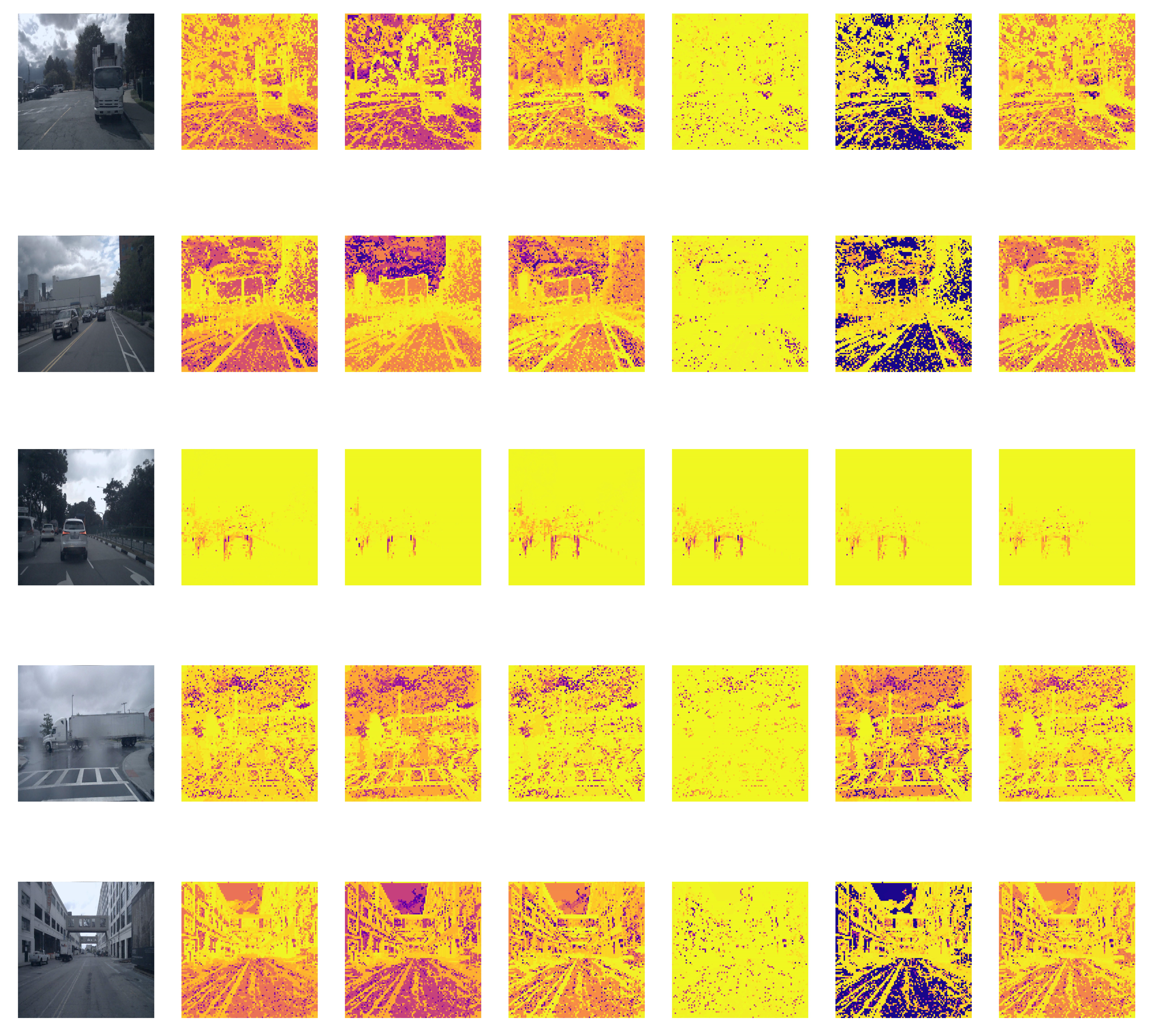} 
 \caption{Raw Images and their Self-Attention Maps Processed by the Context Encoder. The raw images are visualized on the left, while the attention maps of these images in different layers within the encoder are displayed on the right.}
 \label{vit} 
\end{figure*}
\subsubsection{Case Study for the Context Encoder and Cross-modal Attention}
To showcase the effectiveness of the proposed Context Encoder and the cross-modal attention mechanism and their interpretability.
We visualize the prediction results made by CAVG with and without these components in Fig. \ref{failure}. 
Specifically, we select several challenging examples where CAVG, without the Context Encoder and cross-modal attention mechanism, made inaccurate predictions. However, the inclusion of these components markedly improved the predictive accuracy of CAVG, as evidenced by our thorough evaluation.
For a more detailed understanding, Fig. \ref{vit} shows the output of the raw image after processing by the Context Encoder. This processing emphasizes key scene semantics in traffic scenarios, such as lane boundaries, road signs, and pedestrian paths. Moreover, the Context Encoder is able to capture more nuanced environmental cues, such as variations in lighting or indications of weather conditions. 
This extends the visual range of the model beyond the bounding boxes to a more holistic view of the environment. In addition, the cross-modal attention mechanism is instrumental in effectively fusing data from different modalities. Thus, the model goes beyond the visual input and harmonizes it with the given command.

\subsubsection{Subtle Value Analysis of the Emotion Encoder}
Our ablation study, specifically focusing on the Model F without the Emotion Encoder, provides a revealing perspective. At first glance, the empirical improvements attributed to the inclusion of this encoder may seem relatively modest when measured against the broader testset. However, a deeper dive reveals the critical role of this component. We argue that the power of the Emotion Encoder is not captured by broad metrics alone; its true potential emerges in specific, critical scenarios. For example, in a real-world driving context, consider a situation in which a visibly anxious passenger commands the AV to "drive fast"; while the literal interpretation focuses on speed, an understanding of the underlying emotional urgency might prompt the AV to prioritize a destination with fewer obstacles. Such human-friendly responses, attuned to human emotions, could significantly improve passenger comfort. Although these emotion-laden scenarios might constitute a smaller fraction of the entire dataset, their impact is disproportionately significant. 

Importantly, the inclusion of the Emotion Encoder is strategic and lays the groundwork for our future efforts. Grounding linguistic commands in a visual context is only the first step. The next frontier, armed with this understanding of emotional context, is seamless trajectory planning that not only adheres to the explicit command but also resonates with the emotional state of the commander. In essence, by integrating the Emotion Encoder, we expect to evolve CAVG toward truly empathetic and anticipatory autonomous driving systems.

\subsection{Intuition and Interpretability Analysis}
CAVG is constructed with several key design motivations. First, a context encoder captures essential semantic relationships between bounding boxes by ingesting raw images and textual vectors, thereby enriching cross-modal understanding. Second, we present an innovative Emotion Encoder that analyzes the emotional information implicit in commander commands to more deeply capture human-AV interactions.
Furthermore, we introduce a novel cross-modal attention mechanism to harmonize different types of input data. Finally, the RSD layer attention mechanism is employed to fine-tune the fusion of multi-modal vectors and hidden states, enhancing feature interpretability. These combined elements make CAVG both intuitive and adept at making robust predictions in complex scenarios.
\begin{figure*}[htbp]
 \centering \includegraphics[width=0.9\linewidth]{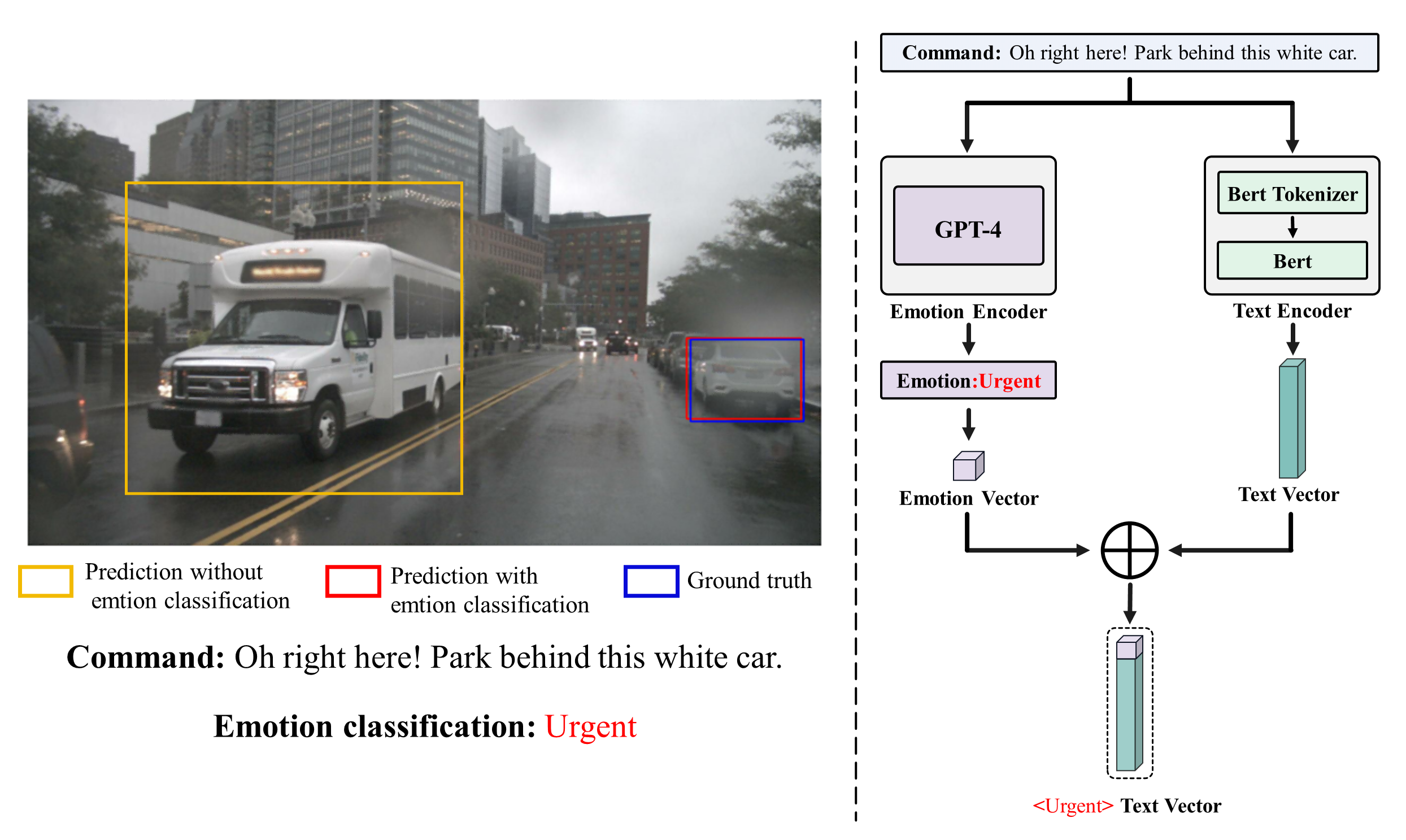} 
 \caption{Illustration of the Emotion Encoder. Demonstrate how the Emotion Encoder indicates the model to identify a specific vehicle using an emotion vector embedded in a text vector.}
 \label{emotion} 
\end{figure*}

Figure \ref{emotion} vividly illustrates how our innovative emotion classification paradigm provides indispensable guidance to CAVG in certain contexts. Given a commander's command ``Oh, right here! Park behind that white car.'', the ambiguity inherent in such a directive becomes clear. Without integrating the commander's emotional context, the model tends to predict parking in front of the approaching white vehicle, resulting in a notable deviation from the ground truth. In contrast, our novel Emotion Encoder delves deep into the linguistic milieu of the command. Through nuanced linguistic analysis, such as recognizing the urgency conveyed by exclamatory phrases or specific keywords, it can identify underlying emotional undertones. In this case, the Emotion Encoder categorizes the emotion as \textbf{``urgent''}. This classification, when fused with the text vector produced by the Text Encoder, produces a richer emotion-sensitive embedding. This fusion communicates to the model the paramount importance and urgency of the commander's request.
With this emotional insight, the model can better understand the commander's immediate needs. This enhanced understanding allows for the mitigation of potential misjudgments caused by ambiguous commands while taking into account human emotions for a more human-centered response.

\begin{figure}[htbp]
 \centering
 \includegraphics[width=0.75\linewidth]{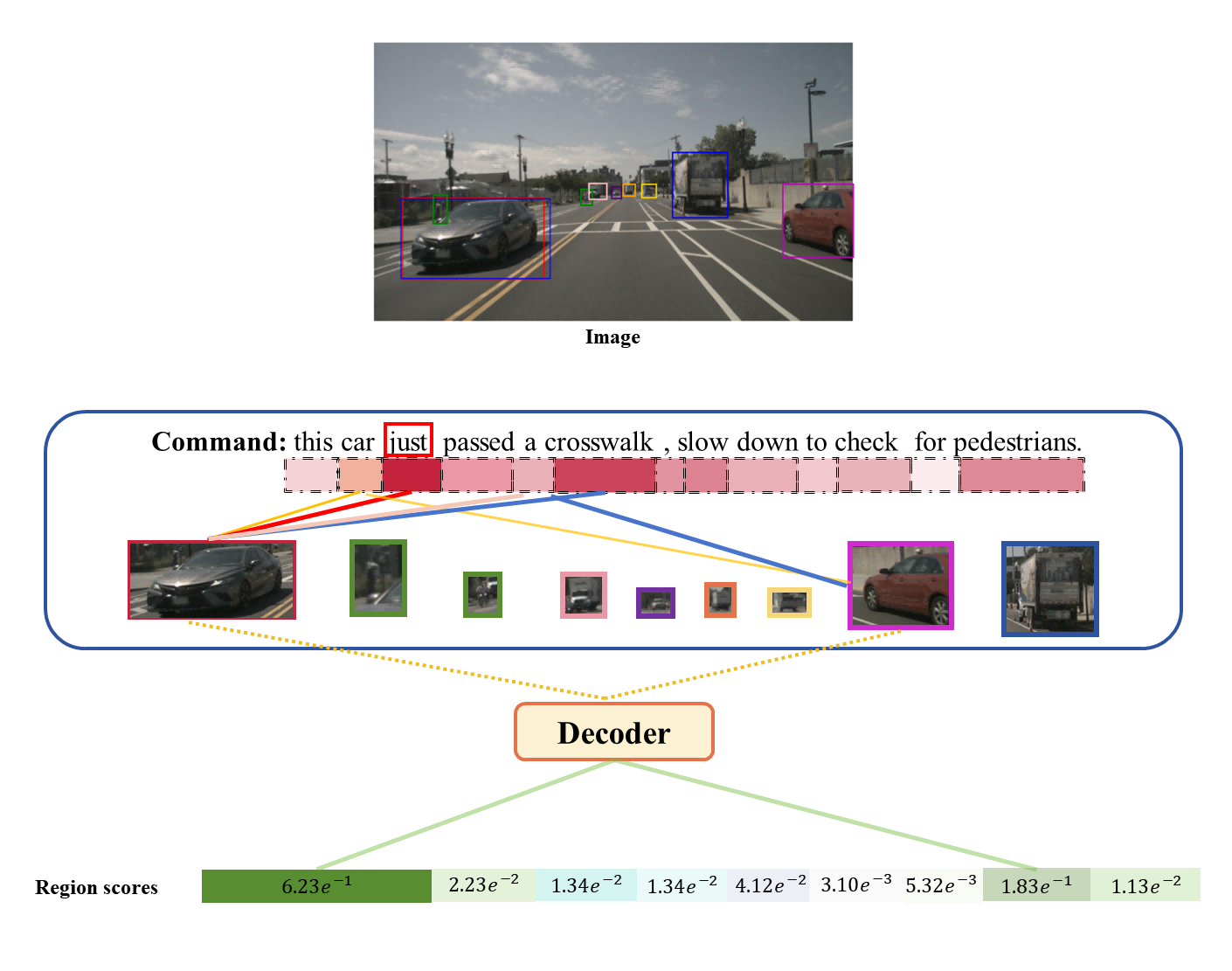}
 \caption{Illustration of the Cross-Modal Attention Mechanism. Demonstration of how keywords such as \textbf{ ``just''} guide the model in identifying specific vehicles, using word vectors and regional features.}
 \label{illustration}
\end{figure}
\begin{figure*}[t]
 \centering \includegraphics[width=\linewidth]{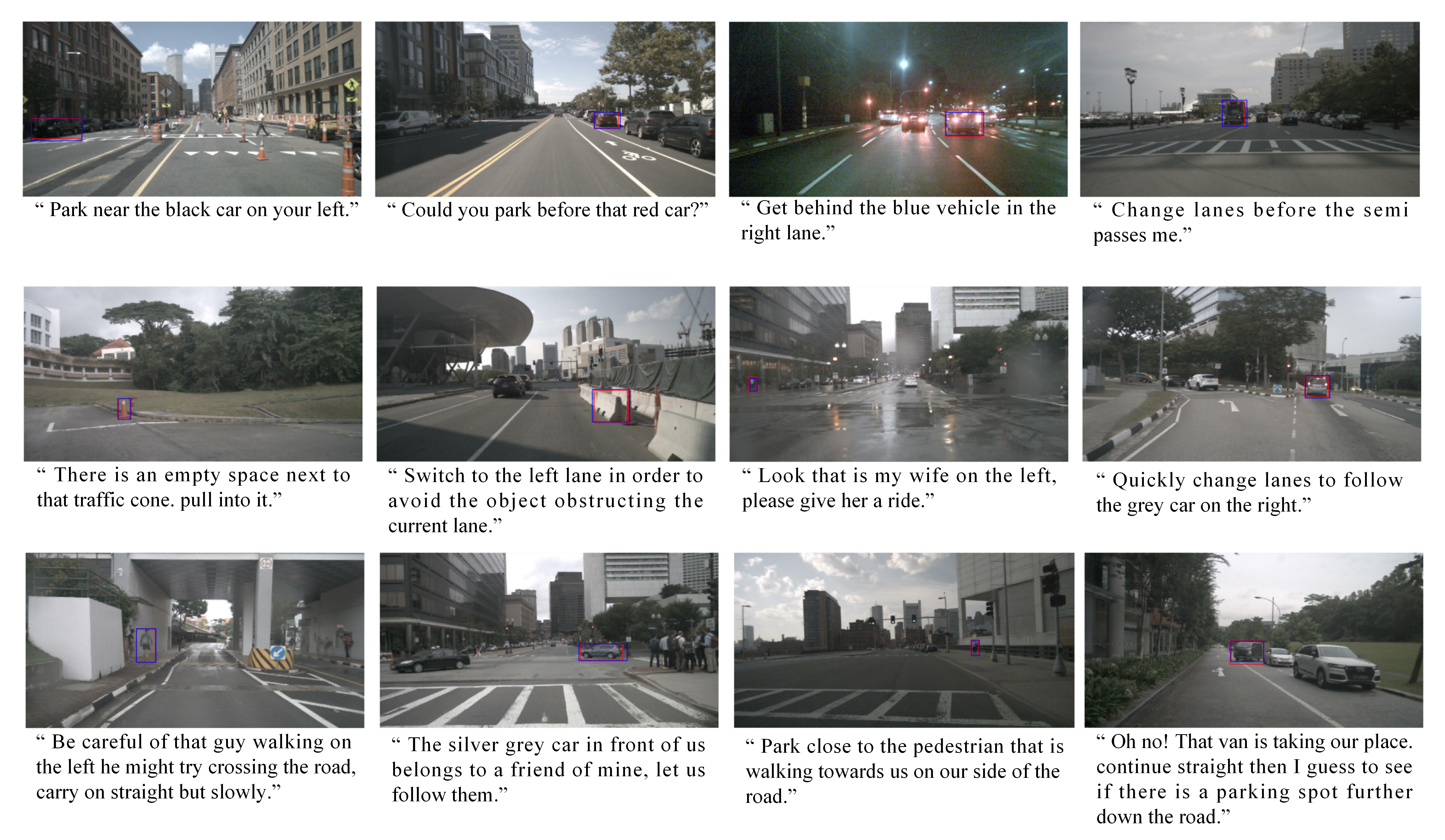} 
 \caption{Comparative Visualization of Model Performance on the Talk2Car Dataset.
  Ground truth bounding boxes are depicted in blue, while output bounding boxes of CAVG are highlighted in red. A natural language command associated with each visual scenario is also displayed below the image for context.}
 \label{qualify} 
\end{figure*}
In addition, in Fig. \ref{illustration}, we demonstrate the crucial role of cross-modal attention mechanisms in disambiguating natural language commands. Specifically, the attention to certain keywords in the command can greatly influence the model's output. For example, in the depicted scenario, the term \textbf{``just''} serves as a key keyword. Overlooking this keyword introduces ambiguity into the system, causing the model to consider multiple vehicles that meet the given criteria. Our multi-modal attention mechanism addresses this by correlating word vectors from the command with regional feature vectors in the image. This allows the model to allocate varying degrees of importance to different segments of the command. For example, only when the model incorporates global features like zebra crossings can it understand that a vehicle has \textbf{``just''} passed the crossing. Consequently, regions that are more semantically aligned with the keywords in the command receive higher attention scores. This enhanced attention mechanism captures not just the features within the bounding boxes but also broader contextual elements, such as lane markings and zebra crossings, allowing the model to make more informed and precise decisions.

\begin{figure*}[htbp]
 \centering \includegraphics[width=\linewidth]{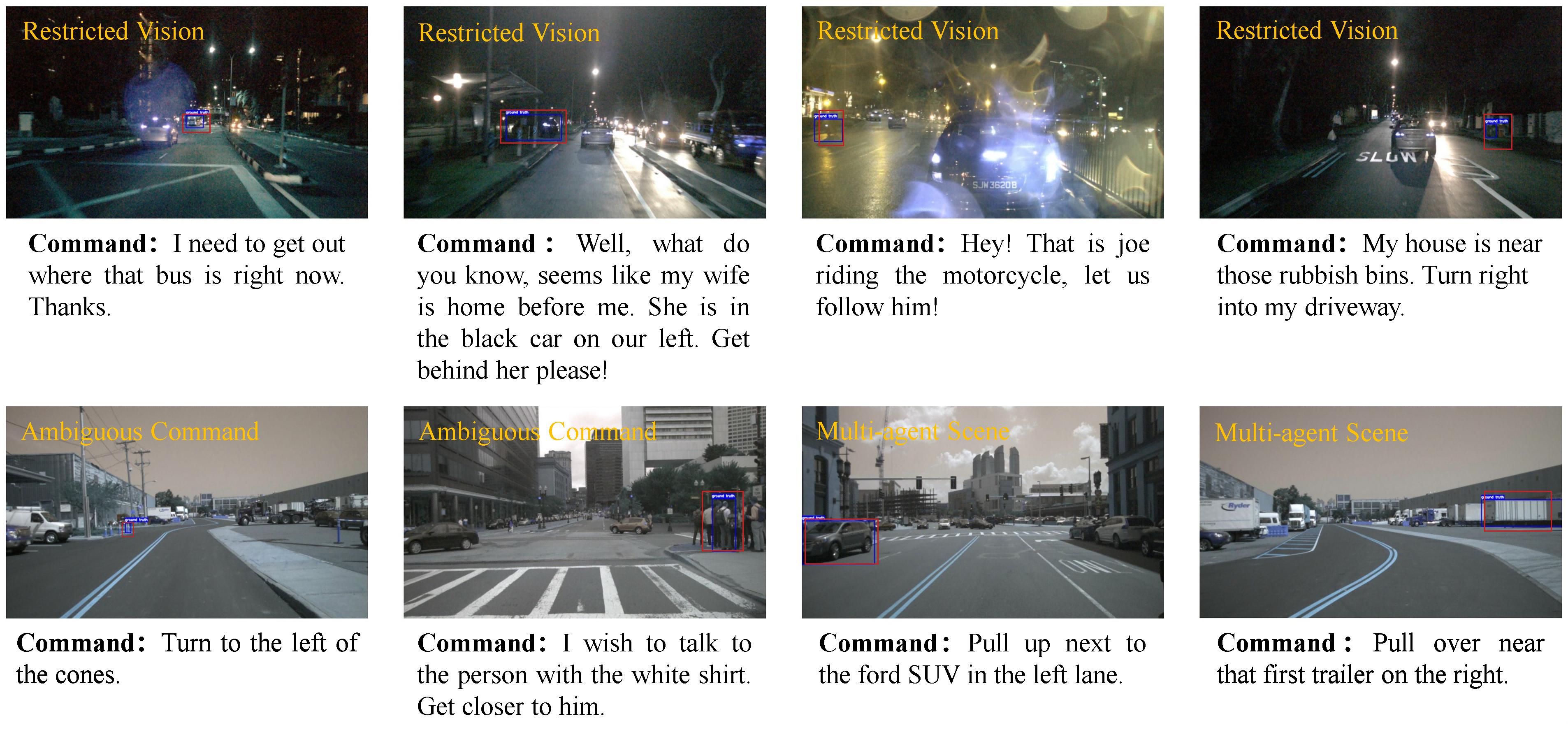} 
 \caption{Comparative Visualization of Model Performance on Challenging Scenes. The challenging scenes include those with limited visibility, ambiguous commands, and scenes with multiple agents. Ground truth bounding boxes are depicted in blue, while output bounding boxes of CAVG are highlighted in red. A natural language command associated with each visual scenario is also displayed below the image for context. }
 \label{qualify_2} 
\end{figure*}

\subsection{Qualitative Results}
To demonstrate the reasoning capabilities of our model, we select a set of challenging driving scenarios for more visual inspection. These selected scenarios exemplify the diverse and multifaceted situations that a vehicle may encounter. 
The visual demonstrations of our model's performance on these instances, derived from the Talk2car dataset, are shown in Fig. \ref{qualify} and Fig. \ref{qualify_2}. CAVG shows remarkable performance in predicting semantic regions from the graphical representation, demonstrating its robustness over a range of environmental conditions. These include challenging conditions such as low-light night scenes, urban environments with complex object interactions, ambiguous command scenes, and rainy days where visibility is compromised.

\section{User Study}\label{User_study}
To assess user satisfaction with our model and SOTA models in real-world applications, we conducted a comprehensive user study using questionnaires. The purpose of this study is to gather user experience and feedback on various visual grounding models for AVs. By analyzing their responses, we sought to evaluate these models in detail and identify their strengths and areas for improvement. This user-centered approach reveals the real-world applicability and effectiveness of these models and contributes to the development of user-friendly AV systems.

\begin{figure*}[htbp]
 \centering \includegraphics[width=0.55\linewidth]{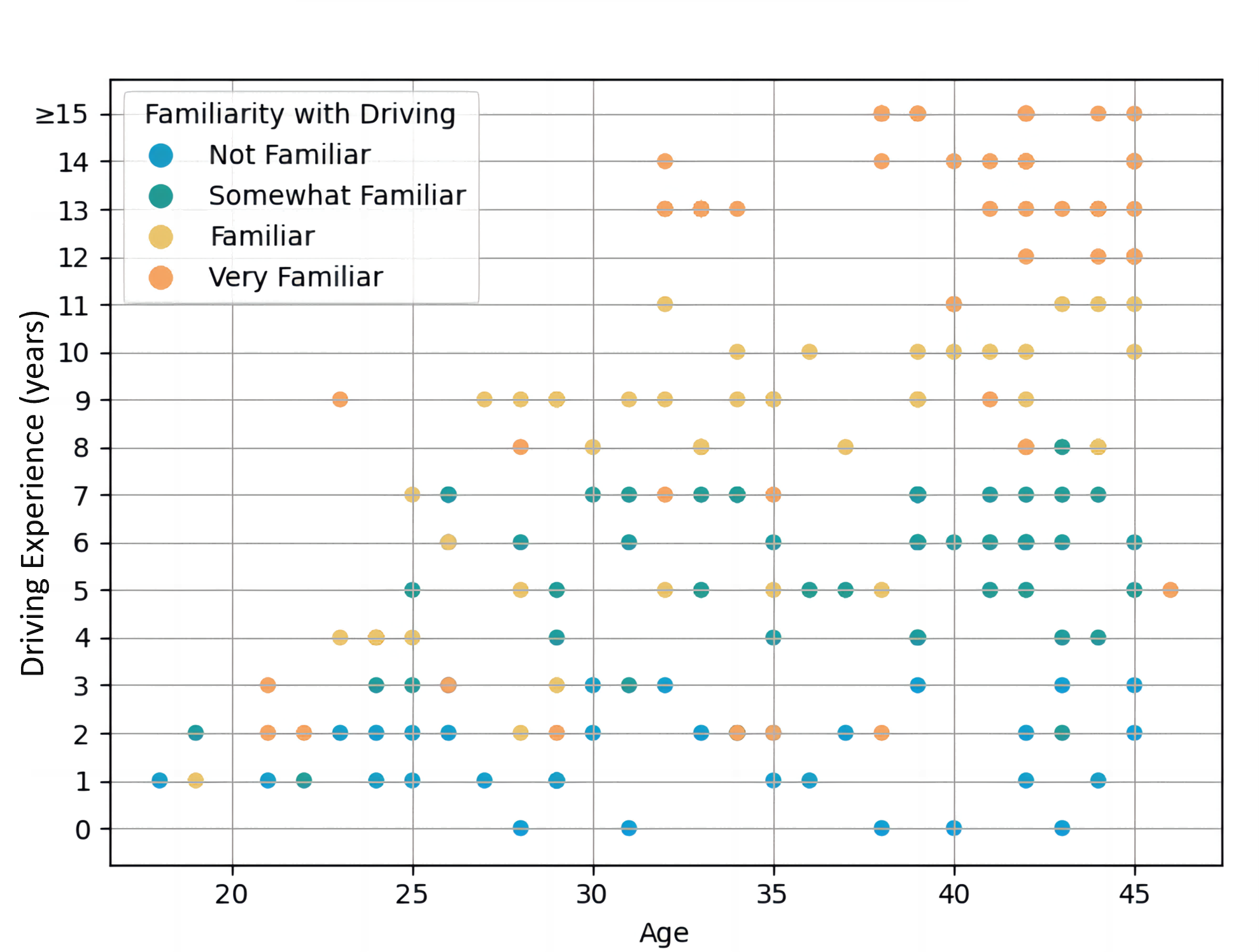} 
 \caption{Age and Driving Experience Distribution of Survey Participants.}
 \label{user_study_1} 
\end{figure*}

Recognizing that user satisfaction is closely related to the driving experience and skills of the participants, our survey included a diverse group of 174 participants, categorized by gender, age, driving experience, and educational background. As shown in Fig. \ref{user_study_1}, the group of participants is evenly distributed and includes users with varying levels of driving experience, including those with extensive experience (over ten years) and those with less experience (one to three years). These participants evaluate five different visual grounding models (CAVG, CAVG (75\%), CAVG (50\%), Stacked VL-BERT, CMSVG, and AttnGrounder) based on three key performance indicators (response accuracy, inference time, and user experience), and select the most preferred model from among them. Notably, to ensure fairness, the survey is anonymous; participants are unaware of the specific categories of the five models. Details of the survey can be found in the Appendix.

\begin{figure*}[t]
 \centering \includegraphics[width=0.95\linewidth]{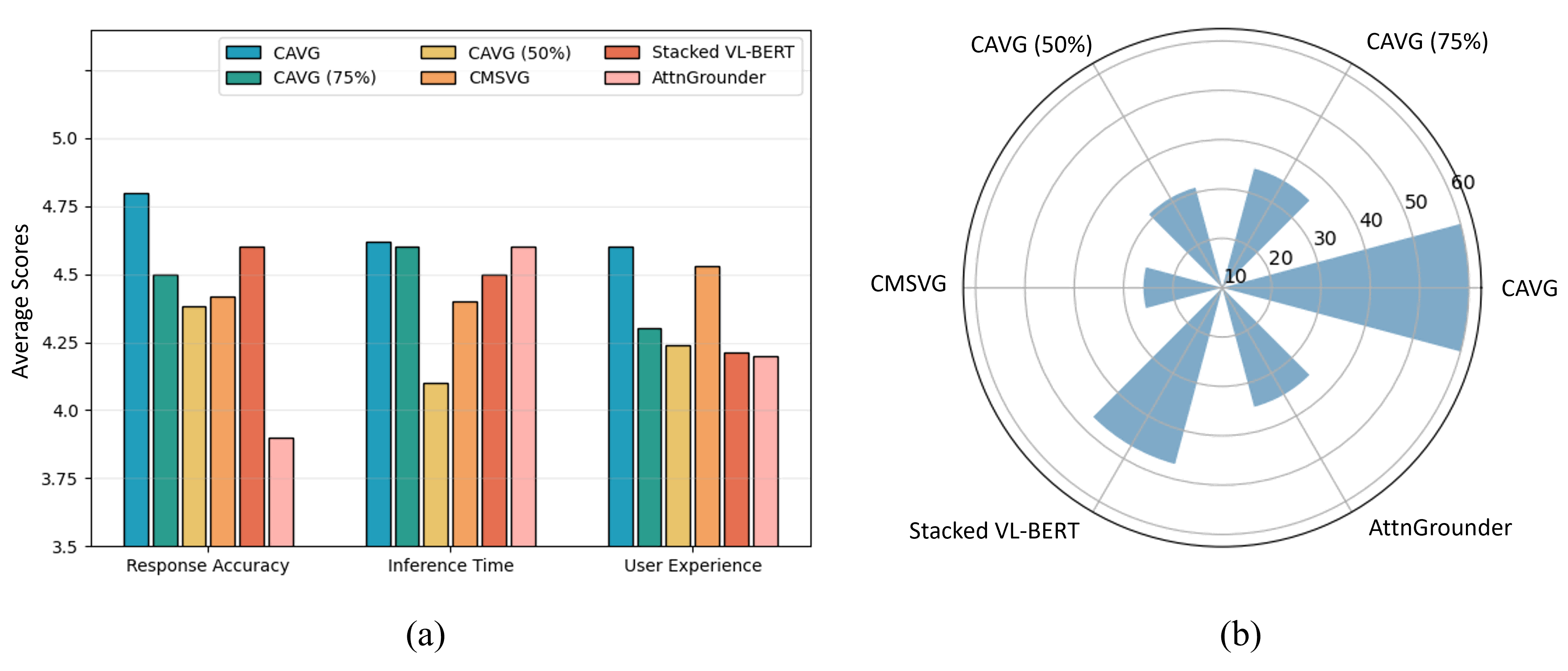} 
 \caption{Evaluation Results for the User Study. (a) Average scores of the models in response accuracy, inference time, and user experience. (b) Quantitative Distribution of favorite models selected by survey participants.}
 \label{user_study_2} 
\end{figure*}

Figure \ref{user_study_2} (a) shows a five-level grading system for the three key user experience metrics, with scores ranging from 0 to 5. CAVG scores highest in terms of response accuracy and user experience alignment, while also maintaining the faster response time. In addition, both the CAVG and CAVG (75\%) provide an impressive user experience and receive relatively high user ratings compared to other models. Overall, CAVG outperforms the existing models by 5.2\% to 10.6\% in terms of comprehensive user ratings. Figure \ref{user_study_2} (b) shows the participants' preferences for the five models. Among the preferred models, CAVG receives the most favor from the participants, indicating CAVG's practicality and significantly higher user satisfaction compared to other models. Crucially, feedback from the survey participants indicates that our future work should focus on further optimizing the structure of the model. This would involve balancing prediction accuracy with improved inference speed to respond more promptly to user commands and meet the real-time demands of visual grounding for autonomous driving, where immediate and accurate response is paramount for safety and user satisfaction.

\section{Conclusion}\label{Conclusion}
In this research, we have developed a CAVG framework, meticulously crafted to enhance visual grounding in AVs. This framework is an integration of five core encoders—Text, Emotion, Vision, Context, and Cross-Modal—each contributing uniquely to the processing of multi-modal data, alongside a Multimodal Decoder. The Text Encoder, utilizing the BERT framework, plays a crucial role in extracting the linguistic nuances within commander commands. It ensures that the subtleties of language are accurately interpreted. Complementing this, our Emotion Encoder, leveraging the capabilities of GPT-4, is pivotal in understanding and classifying the emotional context of commands. This feature enables AVs to respond to human commands in a more empathetic and contextually aware manner. The Vision Encoder is a key component of our model, going beyond conventional object recognition. It employs frameworks like ViT and BLIP to provide a richer contextual understanding of traffic scenes. This capability is crucial for accurately interpreting and responding to grounding commands with a broader range of visual information. Furthermore, the Cross-Modal Encoder introduces a multi-head cross-modal attention mechanism. This mechanism adeptly combines textual and visual information, focusing attention on the most pertinent vectors. The Multimodal Decoder then employs RSD layer attention to fuse insights from each layer, leading to informed multimodal predictions. Tested on the Talk2Car dataset, CAVG demonstrates exceptional prediction accuracy, surpassing existing state-of-the-art methods. Notably, it maintains superior performance even with limited training data, ranging from 50\% to 75\% of the full dataset. Its robustness is further evident under challenging scenarios such as low-light night scenes, long-text command scenarios, rainy days, and dense urban streets.

In conclusion, CAVG not only sets a new benchmark in the AV domain but also significantly enhances the safety and human-centric aspect of the autonomous driving experience. Future research will focus on expanding CAVG's capabilities, including integrating additional modalities like bird's-eye-view cues, lidar feedback, and HD maps. We also plan to investigate CAVG's integration into a comprehensive AV framework, particularly aligning with trajectory planning tasks.

\section*{Acknowledgement}
This research is supported by the Science and Technology Development Fund of Macau SAR (File no. 0021/2022/ITP, 0081/2022/A2, SKL-IoTSC(UM)-2021-2023/ORP/GA08/2022, SKL-IoTSC(UM)-2024-2026/ORP/GA06/2023), and University of Macau (SRG2023-00037-IOTSC). We extend our gratitude to the Institute for Advanced Studies in Humanities and Social Sciences (IAS) for their valuable assistance. For any correspondence regarding this paper, please contact Dr. Zhenning Li at zhenningli@um.edu.mo or Dr. Chengzhong Xu at czxu@um.edu.mo.

\bibliographystyle{unsrt}  
\bibliography{references}

\end{document}